%% file: main.tex
\documentclass{article}

\usepackage[preprint]{corl_2023} 

\usepackage{microtype}
\usepackage{graphicx}
\usepackage{subfigure}
\usepackage{booktabs} 
\usepackage{amsmath}
\usepackage{amssymb}
\usepackage{mathtools}
\usepackage{amsthm}
\usepackage{booktabs}
\usepackage{multirow}
\usepackage{adjustbox}
\usepackage{xcolor}         
\usepackage{xspace}
\usepackage{mathtools}
\usepackage{algorithm}
\usepackage[textsize=tiny]{todonotes}
\usepackage{algorithm}
\usepackage{algorithmic}
\usepackage{wrapfig}

\title{Embodied Executable Policy Learning with Language-based Scene Summarization}

\author{
  Jielin Qiu$^{1,2}$\thanks{Equal contribution}~, Mengdi Xu$^{1,*}$, William Han$^{1,*}$, Seungwhan Moon$^2$, Ding Zhao$^1$ \\
  $^1$Carnegie Mellon University, $^2$Meta Reality Labs \\ 
  {\tt\footnotesize \{jielinq,mengdixu,wjhan,dingzhao\}@andrew.cmu.edu}, 
 {\tt\footnotesize shanemoon@meta.com}
}

\begin{document}
\maketitle


\begin{abstract}
Large Language models (LLMs) have shown remarkable success in assisting robot learning tasks, i.e., complex household planning.
However, the performance of pretrained LLMs heavily relies on domain-specific templated text data, which may be infeasible in real-world robot learning tasks with image-based observations.  Moreover, existing LLMs with text inputs lack the capability to evolve with non-expert interactions with environments.
In this work, we introduce a novel learning paradigm that generates robots’ executable actions in the form of text, derived solely from visual observations, using language-based summarization of these observations as the connecting bridge between both domains. Our proposed paradigm stands apart from previous works, which utilized either language instructions or a combination of language and visual data as inputs. Moreover, our method does not require oracle text summarization of the scene, eliminating the need for human involvement in the learning loop, which makes it more practical for real-world robot learning tasks.
Our proposed paradigm consists of two modules: the SUM module, which interprets the environment using visual observations and produces a text summary of the scene, and the APM module, which generates executable action policies based on the natural language descriptions provided by the SUM module. We demonstrate that our proposed method can employ two fine-tuning strategies, including imitation learning and reinforcement learning approaches, to adapt to the target test tasks effectively.
We conduct extensive experiments involving various SUM/APM model selections, environments, and tasks across 7 house layouts in the VirtualHome environment. Our experimental results demonstrate that our method surpasses existing baselines, confirming the effectiveness of this novel learning paradigm.
\end{abstract}

\keywords{Decision-Making, Embodied Robot Learning, Large Language Model}

\section{Introduction}
\vspace{-10pt}
There has been a surge of interest in building Large Language Models (LLMs) pretrained on large-scale datasets and exploring LLMs' capability in various downstream tasks.
LLMs start from the Transformer model \citep{Vaswani2017AttentionIA} and are first developed to solve different natural language processing (NLP) applications \citep{Devlin2019BERTPO,Liu2019RoBERTaAR,Brown2020LanguageMA}.
Recently, LLMs also show great potential for accelerating learning in many other domains by generating learned embeddings as meaningful representations for downstream tasks and encoding transferable knowledge in large pretraining datasets. 
Examples include transferring the knowledge of LLM to, i.e., 
robotics control \citep{Liang2022CodeAP,Ahn2022DoAI}, 
multimodal learning \citep{Zeng2022SocraticMC,Zellers2021PIGLeTLG},
decision-making \citep{Li2022pretrainedLM,Huang2022LanguageMA},
code generation \citep{Fried2022InCoderAG}, laws \citep{Kaplan2020ScalingLF}, computer vision (CV) \citep{Radford2021LearningTV}, and so on.

In this paper, we focus on the problem of facilitating robot learning by having an LLM in the loop.
The robot generates actions according to its environment observations, which are, in general, sensory information in the format of images, point clouds, or kinematic states. 
We identify one key challenge in massively deploying LLMs to assist robots is that \emph{LLMs lack the capability to understand such non-text-based environment observations}. 
To solve this challenge, \citet{Liang2022CodeAP} utilize rule-based perception APIs to transform image-based observations into text formats, which then serve as inputs to the LLM. 
We instead propose to integrate the multimodal learning paradigm to transform images into texts, which allows more principled and efficient transfer to novel robot learning tasks than rule-based APIs. 
Another key challenge is \emph{the widely-existing large distribution shifts between the training tasks of large pretrained models and testing tasks in the domain of robot learning}. 
To close the domain gap, \citet{Li2022pretrainedLM} adapt the pretrained LLM to downstream tasks via finetuning with observations converted into text descriptions.
In the presence of realistic visual observations, it is still being determined what is an appropriate method to co-adapt pretrained foundation models.

\begin{figure}[t]
  \centering
  \includegraphics[width=0.99\linewidth]{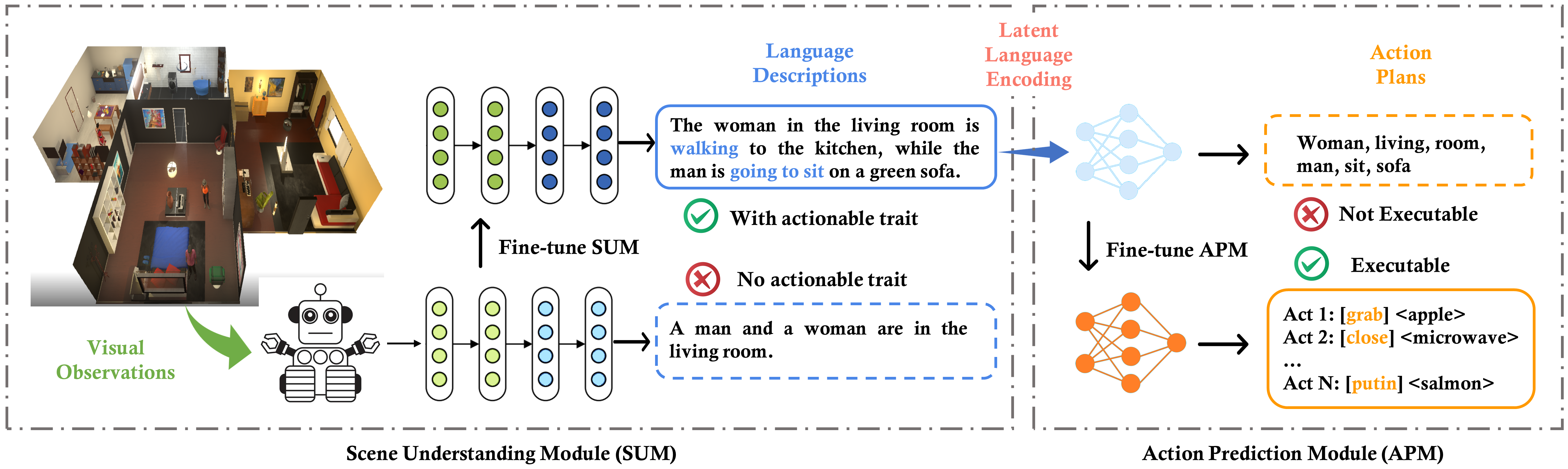}
  \vspace{-5pt}
  \caption{The overall architecture of our approach, which includes a scene understanding module (SUM) and an action prediction module (APM). The agent takes pure visual observations and encodes the information as language. Then the language is transferred to APM for action generation. APM fine-tuned on VirtualHome can generate executable action plans directly.}
  \label{fig:model}
  \vspace{-15pt}
\end{figure}

To address the above challenges, we propose a new visual-based robot learning paradigm that takes advantage of embedded knowledge in both multimodal models and LLMs. 
To align different modalities in the visual observations and text-based actions, we consider language as the bridge information. We build a scene-understanding model (SUM) with a pretrained image captioning model to grant the robot the ability to describe the surrounding environment with natural language.
We then build an action prediction model (APM) with an LLM to generate execution actions according to the scene caption in the format of natural language. 
To adapt pertained models in SUM and APM to downstream robot learning tasks, we propose to finetune the multimodal model in SUM with pre-collected domain-specific image-caption pairs and the language model in APM with corresponding language-action pairs. Besides finetuning with expert demonstrations, we further propose a finetuning paradigm of APM based on the sparse environment feedbacks to endow APM's capability to evolute with non-expert data. 
An illustration of the proposed framework is in Figure~\ref{fig:model}.

Our contributions are summarised as follows:
\vspace{-10pt}
\begin{itemize}
    \item We introduce a novel robot learning paradigm with LLM in the loop that handles multiple modalities of visual observations and text-based actions in a principled manner. We bridge both modalities with natural language generated by a pretrained multimodal model.
    \vspace{-5pt}
    \item To adapt to target testing tasks, we propose two fine-tuning strategies, including imitation learning and reinforcement learning approaches. We collect a new expert dataset for imitation learning-based finetuning.
    \vspace{-5pt}
    \item We test the adaptation performance of multiple models of SUM and APM in seven house layouts in the VirtualHome environment. Our experiments demonstrate that our proposed paradigm shows promising results. Our code is in the Supplementary Material.
\end{itemize}

\vspace{-10pt}
\section{Related Work}
\vspace{-5pt}

\vspace{-5pt}
\paragraph{Language Models for Embodied Agents.}
Recently, several works have successfully combined LLMs with robot learning by taking advantage of the knowledge learned by LLMs, i.e., reasoning \citep{Liang2022CodeAP,Zeng2022SocraticMC,Zellers2021PIGLeTLG}, planning \citep{Shah2022LMNavRN,Huang2022InnerME,Kant2022HousekeepTV,Li2022pretrainedLM,Huang2022LanguageMA}, manipulation \citep{Shafiullah2022CLIPFieldsWS,Jiang2022VIMAGR,Shridhar2022PerceiverActorAM,Bucker2022LaTTeLT,Ren2022LeveragingLF,Tam2022SemanticEF,Khandelwal2022SimpleBE,Shridhar2021CLIPortWA}, and navigation \citep{Lin2022ADAPTVN,Parisi2022TheUE,Gadre2022CLIPOW,Hong2021VLNBERTAR,Majumdar2020ImprovingVN}, which demonstrated the feasibility of using LLM to assist embodied learning. 
In particular, in \citep{Huang2022InnerME}, language instructions are utilized to interpret the scene, whereas we rely on raw image observations. In \citep{singh2022progprompt}, a predefined executable plan prompt is provided without learning from visual observations, simplifying the generation of executable plans. In \citep{xiao2022robotic}, both language instructions and visual images are employed to fine-tune the VLM, which is subsequently used for behavior cloning. However, the generated robot plan consists of high-level natural language instructions rather than executable robot policies, as in our work. In \citep{fan2022minedojo}, the proposed MINECLIP primarily calculates the correlation between an open-vocabulary language goal string and a 16-frame video snippet. The correlation score serves as a learned dense reward function for training a robust multi-task RL agent, which is distinct from our approach.

\vspace{-5pt}
\paragraph{Pretraining and Finetuning Language Models.}

Fine-tuning \citep{Howard2018UniversalLM} has superseded the use of feature extraction of pretrained embeddings \citep{Peters2018DeepCW} while pretrained language models are favored over models trained on many tasks due to their increased sample efficiency and performance \citep{SEBASTIANRecent}. The success of these methods has led to the development of even larger models \citep{Devlin2019BERTPO,Raffel2019ExploringTL}. But those large models may not perform well on data that is different from what they were pretrained on. Under this case, fine-tuning pretrained contextual word embedding models to supervised downstream tasks has become commonplace  \citep{Hendrycks2020pretrainedTI,Dodge2020FineTuningPL}. More related works can be found in Appendix~\ref{sec:appendix-related-work}.

\vspace{-5pt}
\section{Method}
\vspace{-5pt}
In this section, we first introduce our focused problem, which is generating a visual-based policy by leveraging pretrained large models. We then introduce SUM, which learns language descriptions of the surrounding environment, and APM, which predicts actions based on SUM's caption output. To grant both SUM and APM the capability of making the correct understanding and decision in the target domain, we propose their finetuning algorithms. 

\vspace{-5pt}
\paragraph{Problem Formulation.}
\label{sec:sec_problem_formulation}
We consider a general and realistic robot learning task where a robot agent receives a sequential visual observation $V=[v_1, v_2, ..., v_t]$, where $t$ is the timestep, and aims to generate a sequence of actions $A = [a_1, a_2, ..., a_t]$ based on the pure visual observations $V$.
Traditionally, the robot's policy is trained from scratch in the target tasks. Inspired by the success of large pretrained models, we aim to explore the benefit of utilizing pretrained LLMs and multimodal models for general robot learning tasks, where only visual observations are available as inputs. Given the prevailing domain shift between the training domain of the pretrained models and the robot learning tasks, we are motivated to develop a principled finetuning method.

\vspace{-5pt}
\paragraph{SUM: Learning Scene Descriptions from Visual Observations into Language.}
\label{sec:SUM}
The goal of the SUM (scene understanding module) is to transform visual observations into language descriptions that contain an actionable trait to it. SUM shares similar functionalities of visual captioning models, which aim to automatically generate fluent and informative language descriptions of an image \citep{Ke2019ReflectiveDN}. For the SUM to be capable of providing scene descriptions from visual observations, it needs to distill representative and meaningful visual representations from an image, then generate coherent and intelligent language descriptions. 
In our framework, we adopt models with image captioning ability as our SUM. Generally, image captioning models employ a visual understanding system and a language model capable of generating meaningful and syntactically correct captions \citep{Stefanini2021FromST}. In a standard configuration, the task can be defined as an image-to-sequence problem, where the inputs are pixels, which will be encoded as one or
multiple feature vectors in the visual encoding step. Then a language model will take the information to produce a sequence of words or subwords decoded according to a given vocabulary in a  generative way. 

With the development of self-attention \citep{vaswani_2017_attention}, the visual features achieved remarkable performance due to multimodal pretraining and  early-fusion strategies \citep{Tan2019LXMERTLC,Lu2019ViLBERTPT,Li2020OscarOA,Zhou2019UnifiedVP}.
As for language models, the goal is to predict the probability of a given sequence of words occurring in a sentence. As such, it is a crucial component in image captioning, as it gives the
ability to deal with natural language as a stochastic process. Formally, given a sequence of $n$ words $y_1, \ldots, y_n$, the language model component of an image captioning algorithm assigns a probability $P\left(y_1, y_2, \ldots, y_n \mid \boldsymbol{X}\right)$ to the sequence as:
\begin{equation}\small
 P\left(y_1, y_2, \ldots y_n \mid \boldsymbol{X}\right)=\prod^n_{i=1} P\left(y_i \mid y_1, y_2, \ldots, y_{i-1}, \boldsymbol{X}\right),  
\end{equation}
where $\boldsymbol{X}$ represents the visual encoding on which the language model is specifically conditioned. Notably, when predicting the next word given the previous ones, the language model is autoregressive, which means that each predicted word is conditioned on the previous ones. Additionally, the language model decides when to stop generating captions by outputting a special end-of-sequence token.

\vspace{-5pt}
\paragraph{APM: Decoding Language Information into Executable Action Plans.}
\label{sec:APM}
The goal of APM (action prediction module) is to transform latent language information from the SUM output into executable action plans. Since both latent language information and executable action plans are sequential data, an LLM with encoder-decoder architecture is a good option for APM in our framework. In addition, an LLM pretrained on a vast corpus of text already has adequate knowledge, which can be fine-tuned on other tasks to improve learning efficiency. 

An LLM with encoder-decoder architecture suits well for our setting. The encoder is responsible for reading and understanding the input language information from SUM, which is usually based on transformer architecture, and creates a fixed-length vector representation, called the context vector. The decoder then takes the context vector as input and generates the output, in our case, the executable action plans. The decoder uses the context vector to guide its generation of the output and make sure it is coherent and consistent with the input information. However, due to the distribution change between the data that LLM was pretrained on and the new task, the LLM needs to be fine-tuned on the task-specific data to transfer the knowledge. The fine-tuning strategies will be introduced in the following sections.
For LLMs, we use well-adopted pretrained architectures, including BERT \citep{Devlin2019BERTPO}, RoBERTa \citep{Liu2019RoBERTaAR}, and BART \citep{Lewis2020BARTDS}, as both the encoder and decoder.
The goal of the LLM is to learn how to generate programmable, executable actions from the language descriptions outputted by SUM. 

\vspace{-5pt}
\paragraph{Training Pipeline.}

The training pipeline contains two steps. First, we fine-tune SUM with the curated VirtualHome observations (More details about data collection are introduced in Section~\ref{sec:exp_data}). This fine-tuning step is to familiarize SUM with the types of scenes present in the task-specific data. We present pseudocode to fine-tune the SUM in Algorithm~\ref{alg:SUM} in Appendix~\ref{sec:algorithm_code}.

In the second stage, we load the fine-tuned SUM and encode the outputs as latent language embeddings. The embeddings are then fed into the APM, which is then fine-tuned using different fine-tuning loss objectives (supervised one or policy gradient, more details are introduced in Section~\ref{sec:exp}), to achieve the optimal policy with maximum rewards.
The pseudocode for finetuning APM with IL and REINFORCE are in Algorithms~\ref{alg:APM_sup} and \ref{alg:APM_reinforce} in Appendix~\ref{sec:algorithm_code}, respectively.

\vspace{-5pt}
\paragraph{Fine-tuning APM with IL and RL.}

The output word from LLM is sampled from a learned distribution over the vocabulary words. In most simple scenarios, i.e., the greedy decoding mechanism, the word with the highest probability is output. The main drawback of this setting is that possible prediction errors quickly accumulate along the way. To alleviate this drawback, one effective strategy is to use the beam search algorithm \citep{Cho2014LearningPR,Philipp2007} that, instead of outputting the word with maximum probability at each time step, maintaining $k$
sequence candidates and finally outputs the most probable one. For the training or fine-tuning strategies, most strategies are based on  cross-entropy (CE) loss and masked language model (MLM). But recently, RL-based learning objective has also been explored, which  allows  optimizing for captioning-specific non-differentiable metrics directly.

\vspace{-5pt}
\paragraph{Imitation Learning with Cross-Entropy Loss.}

The CE loss aims to minimize the negative log-likelihood of the current word given the previous ground-truth words at each timestep. Given a sequence of target words $y_{1: T}$, the loss is formally defined as:
\vspace{-2pt}
\begin{equation}\small
L_{X E}(\theta)=-\sum_{i=1}^n \log \left(P\left(y_i \mid y_{1: i-1}, \boldsymbol{X}\right)\right),   
\end{equation}
\vspace{-2pt}

where $P$ is the probability distribution induced by LLM, $y_i$ is the ground-truth word at time $i, y_{1: i-1}$ indicate the previous ground-truth words, and $\boldsymbol{X}$ is visual encoding. The cross-entropy loss is designed to operate at the word level and optimize the probability of each word in the ground-truth sequence without considering longer-range dependencies between generated words. Traditional training setting with cross-entropy suffers from exposure bias problems \citep{Ranzato2015SequenceLT} caused by the discrepancy between the training data distribution as opposed to the distribution of its predicted words.

\vspace{-5pt}
\paragraph{Reinforcement Learning with REINFORCE.}

Given the limitations of word-level training strategies observed when using limited amounts of data, a significant improvement was achieved by applying the RL approach. Under this setting, the LLM is considered an agent whose parameters determine a policy. At each time step, the agent executes the policy to choose an action, i.e., the prediction of the next word in the generated sentence. Once the end-of-sequence is reached, the agent receives a reward, and the aim of the training is to optimize the agent parameters to maximize the expected reward \citep{Stefanini2021FromST}.
Similar to \citet{Ranzato2015SequenceLT}, for our policy gradient method, we use REINFORCE \citep{Williams1992SimpleSG,NIPS1999_464d828b}, which uses the full trajectory, making it a Monte-Carlo method, to sample episodes to update the policy.
For fine-tuning LLMs using RL, we frame the problem into an Agent-Environment setting where the agent (policy) can interact with the environment to get the reward for its actions. This reward is then used as feedback to train the model. The mapping of the entities is from the agent (policy), which is an LLM, and the environment (the reward function, also named the model), which generates rewards. The reward function consumes the input as well as the output of the LLM to generate the reward. The reward is then used in a loss function, and the policy is updated. Formally, to compute the loss gradient,
beam search and greedy decoding are leveraged as follows:
\vspace{-2pt}
\begin{equation}\small
\nabla_\theta L(\theta)=-\frac{1}{k} \sum_{i=1}^k\left(\left(r\left(\boldsymbol{w}^i\right)-b\right) \nabla_\theta \log P\left(\boldsymbol{w}^i\right)\right),
\end{equation}
\vspace{-2pt}

where $\boldsymbol{w}^i$ is the $i$-th sentence in the beam or a sampled collection, $r(\cdot)$ is the reward function, and $b$ is the baseline, computed as the reward of the sentence obtained via greedy decoding \citep{Rennie2016SelfCriticalST}, or as the average reward of the beam candidates \citep{Cornia2019MeshedMemoryTF}.
Note that since it would be difficult for a random policy to improve in an acceptable amount of time, the usual procedure entails pretraining with cross-entropy or masked language model first, and then fine-tuning stage with RL by employing a sequence level metric as the reward. This ensures the initial RL policy is more suitable than the random one.

\begin{figure}[t]
\centering
\begin{minipage}[t]{0.56\textwidth}
\centering
 \includegraphics[width=0.8\linewidth]{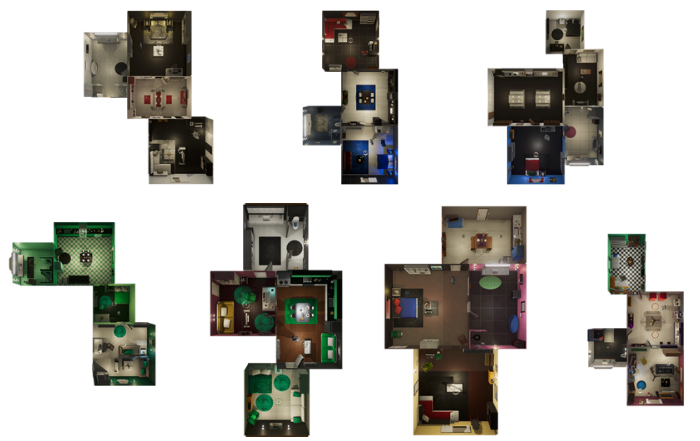}
  \vspace{-5pt}
  \caption{Top-down views of the seven different environments from VirtualHome \citep{Puig2018VirtualHomeSH}.}
  \label{fig:vh_envs}
\end{minipage}
~~
\begin{minipage}[t]{0.4\textwidth}
\centering
\includegraphics[width=0.8\linewidth]{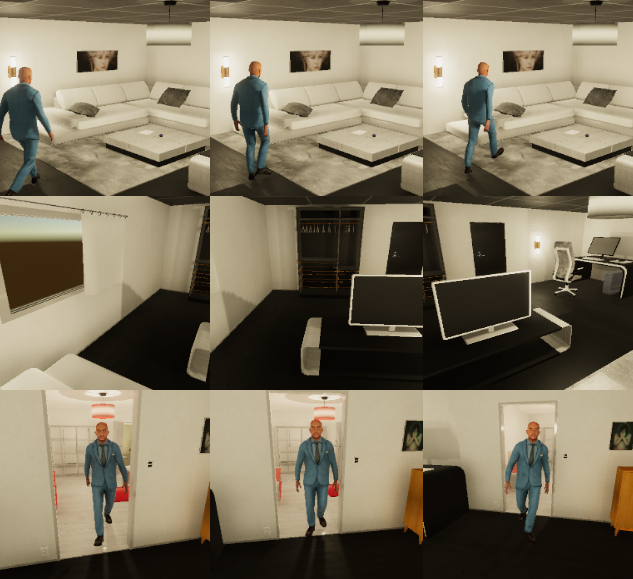}
  \vspace{-5pt}
  \caption{`AUTO', `FIRST\_PERSON', `FRONT\_PERSON' views.}
  \label{fig:vh_data}
\end{minipage}
\vspace{-11pt}
\end{figure}

\begin{table}[t]\small
\caption{Results from different SUMs fine-tuned by the \underline{imitation learning (IL)} objective, where BERT serves as APM. The results are shown on 7 different environments in VirtualHome and also the average performance. The best result in each environment and each SUM model is marked in black and bold. The best SUM result with the highest average performance across 7 environments is marked in \textcolor{orange}{orange} and bold.}
\vspace{-5pt}
\centering
\begin{adjustbox}{width=0.99\linewidth}
\input{tables/diff_models.tex}
\end{adjustbox}
\label{table:SUM_supervised}
\vspace{-15pt}
\end{table}

\vspace{-5pt}
\section{Experiments}\label{sec:exp}

\vspace{-5pt}
This section introduces the environment we used in the experiments, the experimental settings, evaluations, and results. We would like to answer the following questions with experiments: 
(1) Can the proposed paradigm take pure visual observations to generate executable robot actions; 
(2) Which SUMs can provide better scene descriptions for robot learning; 
(3) Which APMs show a better action decoding ability in generating executable actions; 
(4) Which fine-tuning strategies show better performances in this setting; 
(5) Can the model achieve consistent performance across different environments? 

\vspace{-5pt}
\paragraph{Environments.} 
\vspace{-5pt}

We build the experiment environments based on VirtualHome \citep{puig_2018_virtualhome, liao_2019_synthesizing}, a multi-agent, virtual platform for simulating daily household activities. \citep{Puig2018VirtualHomeSH}. 
\citet{puig_2018_virtualhome} provides a dataset of possible tasks in their respective environments. 
Each task includes a natural language description of the task (``Put groceries in the fridge."), an elongated and more detailed natural language description of the task (``I put my groceries into the fridge."), and the executable actions to perform the task in the VirtualHome simulator ([[$Walk$] $<groceries>$ ($1$), [$Grab$] $<groceries>$ ($1$), ... [$Close$] $<fridge>$ ($1$)]).
We define the training and testing tasks based on the natural language descriptions of the task due to their straightforwardness.

In VirtualHome, the agents are represented as 3D humanoid avatars that interact with given environments through provided high-level instructions. 
\citet{puig_2018_virtualhome} accumulated a knowledge base of instructions by using human annotators from AMT to first yield verbal descriptions of verbal activities. These descriptions were further translated by AMT annotators into programs utilizing a graphical programming language, thus amassing around 3,000 household activities in 50 different environments \citep{puig_2018_virtualhome}.
In this study, we evaluate our model's performance in seven unique environments in VirtualHome, which are shown in Figure~\ref{fig:vh_envs}. Each environment has a distinctive set of objects and actions that may be interacted with by agents.

\vspace{-5pt}
\paragraph{Metrics.} 

We used standard NLP evaluation metrics, i.e., BLEU \citep{Papineni2002BleuAM}, ROUGE \citep{Lin2004ROUGEAP}, METEOR \citep{Banerjee2005METEORAA}, CIDEr \citep{vedantam_2015_cider}, and SPICE \citep{DBLP:journals/corr/AndersonFJG16}, for evaluating LLMs.
In addition, we introduced the execution rate following \citet{Li2022pretrainedLM}. The execution rate is defined as the probability of the agent's success in performing the outputted action from APM over the whole trajectory. More experimental setup details about SUM and APM are listed in Appendix~\ref{sec:appendix-experimental-setup}.

\vspace{-5pt}
\paragraph{Datasets.}\label{sec:exp_data}

To fine-tune SUM and APM on task-specific robot learning scenarios, we collect data via VirtualHome, including the agent's observations, language instructions, and action sequences. During data collection, a household activity program can be described as: [[$action_i$] $<object_{i}>$ ($id_{i}$), ... [$action_n$] $<object_{n}>$ ($id_{n}$)], where $i$ denotes each step of the program, $action_i$ and $object_i$ denotes the action performed on the object at step $i$, and $id_i$ symbolizes the unique identifier of $object_i$ \cite{puig_2018_virtualhome}.
The original dataset was augmented by ResActGraph \citep{liao_2019_synthesizing}.
After augmentation, the dataset contains over 30,000 executable programs, with each environment containing over 300 objects and 4,000 spatial relations.
Additionally, we collect the image and text pairs separated by the environments they were executed in. This is important due to the different objects and actions available in each environment.
However, as noted in \citet{puig_2018_virtualhome} and \citet{liao_2019_synthesizing}, not all programs were executable.

During data collection, we observed that text descriptions often comprise just two words (i.e., walk bathroom, sitting chair, run treadmill). To have a more robust description, we prompt-engineered the text with a fill-mask pipeline using BERT \citep{Devlin2019BERTPO, song_2019_mass}. 
For this study, we collect programs executed in three different views: `AUTO', `FIRST\_PERSON', and `FRONT\_PERSON' as shown in Figure~\ref{fig:vh_data}. 
In the `AUTO' view, there are locked cameras in every scene through which the program randomly iterates. The `FIRST\_PERSON' view observes the agent's actions through the first-person point of view. The `FRONT\_PERSON' view monitors the agent's actions through the front in a locked third-person point of view. 
Therefore, the final count of image-text pairs for our dataset in the `AUTO', `FIRST\_PERSON', and `FRONT\_PERSON' views are 26,600, 26,607, and 26,608, respectively.

\vspace{-10pt}
\section{Results and Discussions}
\vspace{-5pt}
\subsection{Model Performance with IL Fine-tuning}

\begin{wrapfigure}{r}{5.9cm}
\vspace{-25pt}
  \centering
  \includegraphics[width=0.8\linewidth]{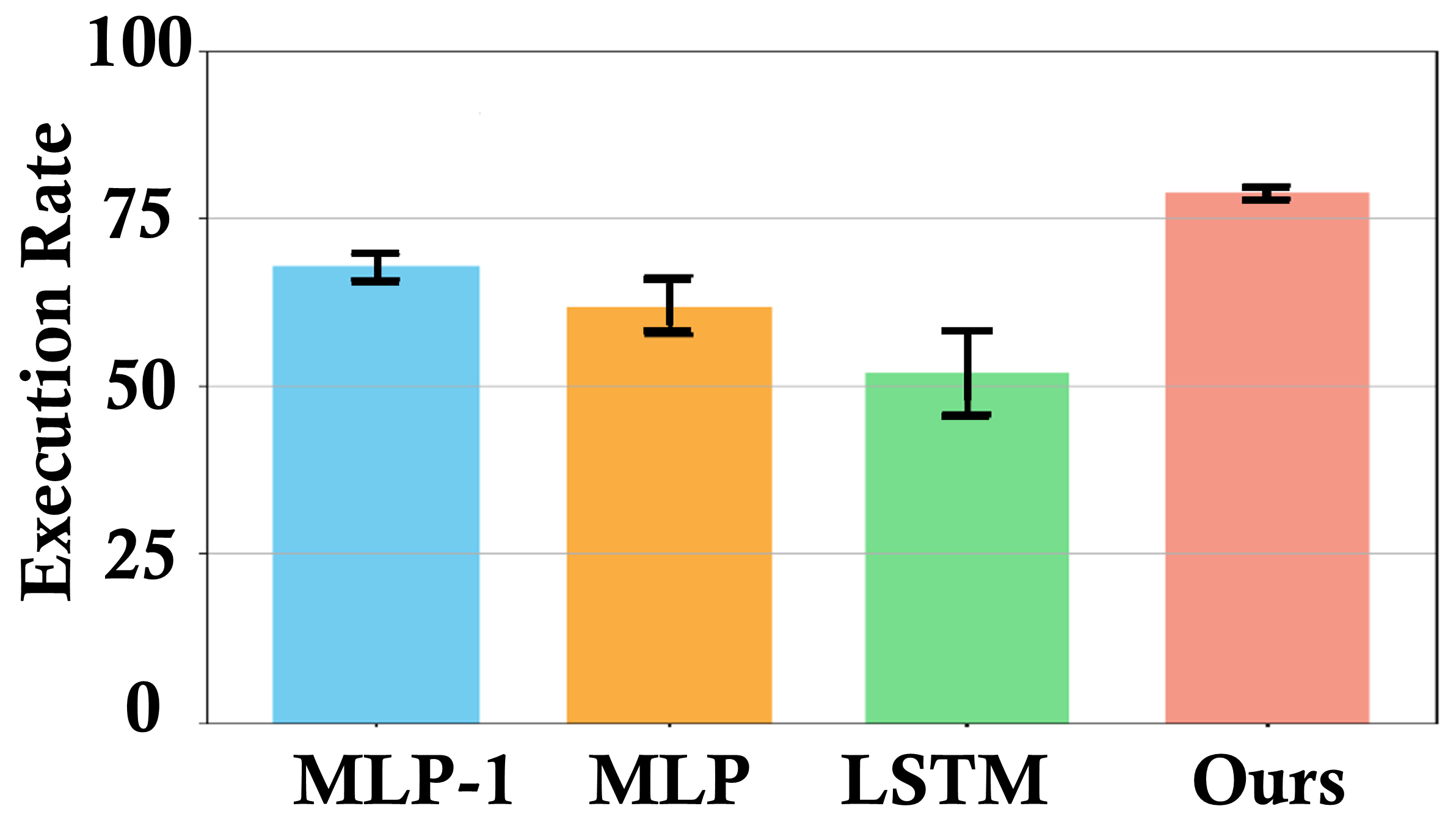}
  \caption{Comparison of our approach with baseline methods in the \underline{imitation learning} setting
   evaluated by the execution rate. }
  \label{fig:comparison}
  \vspace{-8pt}
\end{wrapfigure}
\vspace{-5pt}
We first want to show the benefit of the proposed framework compared with other model architectures. Concretely, in the IL setting with expert data, we compare the execution rate of our model with the {\fontfamily{pcr}\selectfont MLP}, {\fontfamily{pcr}\selectfont MLP-1} and {\fontfamily{pcr}\selectfont LSTM} baselines in \citet{Li2022pretrainedLM}.
Our model has OFA in SUM and BART as APM.
Note that all the baselines are not trained by datasets in other domains and have structured text input instead of realistic visual inputs as our proposed model.
In the {\fontfamily{pcr}\selectfont LSTM} baseline, the hidden representation from the last timestep, together with the goal and current observation, are used to predict the next action. {\fontfamily{pcr}\selectfont MLP} and {\fontfamily{pcr}\selectfont MLP-1} both take the goal, histories, and the current observation as input and send them to MLPs to predict actions. {\fontfamily{pcr}\selectfont MLP-1} has three more average-pooling layers than {\fontfamily{pcr}\selectfont MLP} that average the features of tokens in the goal, history actions, and the current observation, respectively, before sending them to the MLP layer. More details about the baselines can be found in \citet{Li2022pretrainedLM}. As shown in Figure~\ref{fig:comparison}, our approach outperforms baselines in \citet{Li2022pretrainedLM} in terms of a higher average execution rate and a smaller standard deviation, though all the methods are trained on expert data with imitation learning objectives. 
The results show that the pretrained embeddings and large model architecture benefit the performance in downstream robot learning tasks.

\vspace{-5pt}
\subsection{Model Performance with RL Fine-tuning}

\begin{wraptable}{r}{4.9cm}\small
\vspace{-20pt}
\caption{Results from different APMs fine-tuned by the \underline{REINFORCE} loss objective. The results are shown by the average of 7 different environments in VirtualHome. The best results are marked in bold.}
\vspace{5pt}
\centering
\begin{adjustbox}{width=0.95\linewidth}
\input{tables/reinforce_llm.tex}
\end{adjustbox}
\label{table:APM_reinforce}
\end{wraptable}
We provide the model performance after fine-tuning SUM with a frozen BERT in Table~\ref{table:SUM_supervised} for the IL setting with expert data and in Table~\ref{table:SUM_reinforce} for the RL setting. 
We further provide the performance after fine-tuning APM with the fine-tuned SUM in Table~\ref{table:APM_supervised} and Table~\ref{table:APM_reinforce}.
We can see that fine-tuning with expert data in IL results in higher average and per-environment performance than fine-tuning with RL, which shows the benefit of having access to the expert datasets. However, fine-tuning with RL still brings performance improvement to 57.2\% as in Table~\ref{table:APM_reinforce}. Note that without finetuning, the outputs of the LLMs in APM are generally not executable, as shown in Figure~\ref{fig:model}.
Moreover, we consistently observe that the combination of having OFA in SUM and BART as APM achieves the best performance after both IL (Table~\ref{table:APM_supervised}) and RL (Table~\ref{table:APM_reinforce}) fine-tuning.

\vspace{-5pt}
\subsection{Ablation Study}
\vspace{-5pt}
To better understand the importance of different components in our paradigm that affect the overall performance, we conduct ablation studies on different factors, including varying components in SUM, components in APM, and environmental variations.

\vspace{-5pt}
\paragraph{Different Components in SUM.}
The performances of using different components in SUM for IL and RL fine-tuning are in Table~\ref{table:SUM_supervised} and Table~\ref{table:SUM_reinforce}, respectively. From Table~\ref{table:SUM_supervised}, we see that with expert data, OFA achieves better results than BLIP and GRIT on the average performance over 7 environments.
We conjecture that this may be due to OFA being pretrained on 20M image-text pairs, which is larger than the size of other models' pretraining data. While under REINFORCE fine-tuning loss, as in Table~\ref{table:SUM_reinforce}, BLIP slightly outperforms OFA in terms of average performance but has around 4 times larger standard deviation than OFA.

\begin{table}[t]\small
\caption{Results from different APMs fine-tuned by the \underline{imitation learning (IL)} loss objective. The results are shown by the average of 7 different environments in VirtualHome. The results are shown in \%, and the best results are marked in bold.}
\centering
\begin{adjustbox}{width=0.99\linewidth}
\input{tables/diff_LLM.tex}
\end{adjustbox}
\label{table:APM_supervised}
\vspace{-5pt}
\end{table}

\begin{table}[t]\small
\caption{Execution Rates by different SUM fine-tuned by \underline{REINFORCE}, where BERT serves as APM. The results are shown on 7 different environments in VirtualHome and also the average performance. The best results are marked in bold.}
\centering
\begin{adjustbox}{width=0.99\linewidth}
\input{tables/reinforce_sum.tex}
\end{adjustbox}
\label{table:SUM_reinforce}
\vspace{-15pt}
\end{table}

\vspace{-5pt}
\paragraph{Different Components in APM.}
The results of using different components in APM for IL and RL fine-tuning are presented in Table~\ref{table:APM_supervised} and Table~\ref{table:APM_reinforce}, respectively.
We found that BART consistently outperforms other LLMs in both settings.
We hypothesize that due to BART's architectural nature as a denoising autoencoder, it is more suitable for translating natural language descriptions into executable action programs for the VirtualHome simulator.

\vspace{-5pt}
\paragraph{Different Environments.}

To test the performance variations under different environments, we conducted the experiments separately for each unique environment.
The results are shown in Table~\ref{table:SUM_supervised} and Table~\ref{table:SUM_reinforce}, for fine-tuning SUM under IL and RL settings, respectively. Due to image observation variations having the most impact on SUM instead of APM, so we only test the performance of SUM under different environment settings. Through Table~\ref{table:SUM_supervised} and Table~\ref{table:SUM_reinforce}, we could find that the variations exist among different environments. Generally, environment 6 seems to have the easiest environmental settings for the model to learn.

\vspace{-5pt}
\paragraph{Stability.} To evaluate the stability of different models under different envs, we calculated the standard deviation of the results across different trials. The results are shown in Tables~\ref{table:SUM_supervised},\ref{table:APM_reinforce},\ref{table:APM_supervised},\ref{table:SUM_reinforce}, which shows that BART as APM and OFA as SUM seems more stable than the rest of the combinations.

\vspace{-5pt}
\section{Conclusion}
\vspace{-5pt}
In this work, we introduce a novel robot learning paradigm with LLM in the loop that handles multiple modalities of visual observations and text-based actions in a principled manner. We bridge both modalities with natural language generated by a pretrained multimodal model.
Our model contains SUM and APM, where SUM uses image observations as inputs taken by the robot to generate language descriptions of the current scene, and APM predicts the corresponding actions for the next step. We tested our method in the VirtualHome under 7 unique environments, and the results demonstrated that our proposed paradigm outperforms the baselines in terms of execution rates and shows strong stability across environments. 

\vspace{-5pt}
\paragraph{Limitations.}

(1) In our current study, we primarily focused on abstract high-level actions represented by language commands, without taking into account low-level controls such as joint motor control. This omission of the low-level control module may limit the overall effectiveness of the learned policies and their ability to function in complex and dynamic environments. An interesting future direction would be to consider the physical capabilities of embodied agents by learning universal low-level controllers for various morphologies.
(2) Our study might encounter challenges related to long-tailed actions. In our collected dataset, there are actions that occur infrequently, and the current method may not have effectively learned policies for scenarios involving such actions that rarely appear in the collected dataset. This limitation could constrain the overall effectiveness of the learned policies in real-world situations.
(3) Given that we fine-tuned the model using a dataset collected in the VirtualHome environment, the generalizability of the learned policies to other platforms might be insufficient due to significant differences between various simulated platforms.

\clearpage
\bibliography{egbib} 

\newpage
\appendix
\onecolumn
\input{appendix.tex}

\end{document}

%% file: tables/diff_models.tex
\begin{tabular}{lcccccccccc}
\toprule
SUM/Results(\%) & Environment  & Bleu-1 & Bleu-2 & Bleu-3 & Bleu-4 & ROUGE-L & METEOR & CIDEr & SPICE&Execution Rate\\
\midrule
\multirow{8}{*}{OFA}
&1 & 55.1$\pm0.05$ & 45.4$\pm0.10$ & 36.5$\pm0.20$ & 23.0$\pm0.00$ & 60.0$\pm0.16$ & 33.4$\pm0.00$ & 30.2$\pm0.44$ & \textbf{49.9}$\pm0.43$ & 78.0$\pm2.39$\\
&2 & 58.0$\pm0.20$ & 41.7$\pm0.19$ & 35.1$\pm1.01$ & 22.1$\pm0.73$ & 60.1$\pm0.50$ & 34.1$\pm0.52$ & 30.3$\pm0.71$ & 48.1$\pm0.41$ & 79.9$\pm2.37$\\
&3 & 55.3$\pm0.30$ & 42.3$\pm0.62$ & 34.9$\pm0.15$ & 23.0$\pm0.00$ & 60.5$\pm0.01$ & 34.8$\pm0.64$ & 31.2$\pm0.55$ & 48.4$\pm0.17$ & 80.0$\pm3.29$\\
&4 & 57.8$\pm0.73$ & 42.2$\pm0.31$ & 35.3$\pm0.38$ & 24.5$\pm0.67$ & 59.9$\pm0.45$ & 34.6$\pm0.54$ & 33.1$\pm0.63$ & 49.0$\pm0.66$ & 79.9$\pm4.14$\\
&5 & 59.4$\pm0.44$ & 40.3$\pm0.03$ & 34.8$\pm0.02$ & 24.2$\pm0.37$ & 59.7$\pm0.25$ & 35.1$\pm0.62$ & 32.7$\pm0.24$ & 38.0$\pm0.13$ & 77.4$\pm1.12$\\
&6 & \textbf{60.5}$\pm0.01$ & \textbf{48.1}$\pm0.53$ & \textbf{36.6}$\pm0.07$ & \textbf{25.1}$\pm0.15$ & \textbf{61.9}$\pm0.13$ & \textbf{36.2}$\pm0.60$ & \textbf{34.6}$\pm1.07$ & \textbf{49.9}$\pm0.77$ & \textbf{80.5}$\pm1.13$\\
.&7 & 58.2$\pm0.30$ & 46.5$\pm0.58$ & 34.6$\pm0.04$ & 22.3$\pm0.08$ & 58.3$\pm0.92$ & 35.6$\pm0.62$ & 30.8$\pm0.37$ & 44.2$\pm0.33$ & 69.2$\pm2.31$\\
&Average & \textcolor{orange}{\textbf{57.8$\pm0.92$}} & \textcolor{orange}{\textbf{43.8$\pm1.02$}} & \textcolor{orange}{\textbf{35.4$\pm0.63$}} & \textcolor{orange}{\textbf{23.5$\pm0.77$}} & 60.1$\pm0.41$ & \textcolor{orange}{\textbf{34.8$\pm0.62$}} & \textcolor{orange}{\textbf{31.8$\pm1.31$}} & \textcolor{orange}{\textbf{46.8$\pm0.80$}} & \textcolor{orange}{\textbf{77.8$\pm3.26$}}\\
\midrule
\multirow{8}{*}{BLIP}
& 1 &51.1$\pm0.50$ &42.6$\pm0.41$ & 33.2$\pm0.34$ & 21.1$\pm$0.63 & 60.8$\pm0.73$ & 34.7$\pm0.63$ & \textbf{35.5}$\pm00.09$ & 42.7$\pm0.91$ &72.6$\pm1.99$ \\
&2 & 50.5$\pm0.87$ & 41.8$\pm0.72$ & 30.5$\pm28$ & 22.3$\pm0.34$ & 60.3$\pm0.64$ & 33.6$\pm0.87$ & 30.0$\pm0.72$ & 42.8$\pm0.99$ &66.1$\pm4.21$\\
&3 & 52.4$\pm0.54$ & 43.2$\pm0.65$ & 33.6$\pm0.13$ & 21.1$\pm0.52$ & 61.4$\pm0.29$ & 34.5$\pm0.12$ & 31.1$\pm0.00$ & \textbf{48.9}$\pm0.80$&85.0$\pm3.32$\\
&4 & 51.0$\pm1.19$ & 42.1$\pm0.87$ & \textbf{33.8}$\pm0.54$ & 22.8$\pm0.65$ & 60.6$\pm0.76$ & 34.4$\pm0.98$ & 35.1$\pm0.85$ & 46.0$\pm0.74$&73.0$\pm3.65$\\
&5 & 49.0$\pm0.53$ & 38.8$\pm0.43$ & 30.4$\pm0.72$ & 20.0$\pm0.47$ & 58.6$\pm0.65$ & 34.1$\pm0.75$ & 21.0$\pm0.66$ & 30.8$\pm0.69$&67.2$\pm0.93$\\
&6 & \textbf{52.6}$\pm0.79$ & \textbf{44.5}$\pm0.00$ & 31.0$\pm0.63$ & \textbf{24.8}$\pm0.62$ & \textbf{62.0}$\pm0.73$ & \textbf{35.3}$\pm1.02$ & 31.0$\pm0.02$ & 42.4$\pm0.87$&84.1$\pm3.54$\\
&7 & 52.7$\pm0.50$ & 44.0$\pm0.21$ & 33.6$\pm0.18$ & 24.0$\pm0.52$ & 61.7$\pm0.08$ & 34.5$\pm0.60$ & 34.5$\pm0.81$ & 48.8$\pm0.28$&\textbf{86.0}$\pm4.92$\\
&Average & 51.3$\pm0.31$ & 42.4$\pm0.54$ & 32.3$\pm0.66$ &22.3$\pm0.31$& 60.7$\pm0.63$ & 34.4$\pm0.75$ & 31.2$\pm0.87$ & 43.2$\pm0.97$&76.3$\pm5.22$\\
\midrule
\multirow{8}{*}{GRIT} 
& 1 & 50.5$\pm0.99$ & 40.5$\pm0.86$ & 31.8$\pm1.82$ & 20.7$\pm1.02$ & 60.0$\pm1.44$ & 33.1$\pm0.97$ & 30.4$\pm1.42$ & 41.7$\pm0.85$ & 69.2$\pm5.57$  \\
&2 & 52.1$\pm0.66$ & 41.8$\pm1.77$ & 31.7$\pm1.92$ & 20.1$\pm0.97$ & 59.9$\pm0.65$ & 32.1$\pm0.76$ & 29.4$\pm0.87$ & 42.0$\pm0.88$ & 71.4$\pm5.52$\\
&3 & 52.3$\pm0.88$ & 40.3$\pm0.82$ & 32.1$\pm0.77$ & 19.9$\pm1.53$ & 60.4$\pm0.68$ & 31.7$\pm0.66$ & 30.1$\pm2.52$ & 43.5$\pm1.64$ & 71.3$\pm5.98$\\
&4 & 51.9$\pm0.93$ & 39.8$\pm0.92$ & 31.8$\pm0.97$ & 21.3$\pm1.72$ & 59.7$\pm1.22$ & 32.0$\pm0.76$ & 30.0$\pm0.79$ & 42.8$\pm0.84$ & 72.8$\pm4.65$\\
&5 & \textbf{54.7}$\pm0.93$ & 42.3$\pm1.02$ & 33.2$\pm1.25$ & 24.5$\pm0.93$ & 62.3$\pm1.42$ & \textbf{33.8}$\pm1.77$ & 30.7$\pm1.32$ & \textbf{44.6}$\pm1.23$ & \textbf{78.5}$\pm5.07$\\
&6 & 54.6$\pm1.42$ & \textbf{44.7}$\pm1.64$ & \textbf{34.1}$\pm1.32$ & \textbf{25.8}$\pm1.22$ & \textbf{65.8}$\pm1.25$ & 30.1$\pm2.31$ & \textbf{34.5}$\pm0.72$ & 44.0$\pm0.96$ & 78.4$\pm3.66$\\
&7 & 53.9$\pm0.88$ & 42.0$\pm1.79$ & 32.6$\pm2.00$ & 22.5$\pm0.90$ & 63.4$\pm1.00$ & 31.8$\pm1.23$ & 32.3$\pm1.31$ & 43.1$\pm1.41$ & 70.0$\pm3.99$\\
&Average & 52.9$\pm0.18$ & 41.6$\pm0.87$ & 32.4$\pm0.72$ & 22.1$\pm0.68$ & \textcolor{orange}{\textbf{61.6$\pm0.53$}} & 32.1$\pm0.33$ & 31.1$\pm0.25$ & 43.1$\pm0.76$ & 73.1$\pm3.11$\\
\bottomrule
\end{tabular}

%% file: tables/reinforce_llm.tex
\begin{tabular}{lcc}
\toprule
APM& SUM &Execution Rate (\%)\\
\midrule
\multirow{3}{*}{BERT}
&OFA & \textbf{54.7}$\pm1.15$\\
&BLIP & 54.1$\pm1.37$\\
&GRIT & 53.9$\pm3.00$\\
\midrule
\multirow{3}{*}{RoBERTa}
& OFA &  \textbf{55.6}$\pm4.31$\\
& BLIP & 55.2$\pm1.16$\\
& GRIT & 54.8$\pm2.54$\\
\midrule
\multirow{3}{*}{BART}
& OFA &\textbf{57.2}$\pm2.43$\\
& BLIP & 57.0$\pm3.12$\\
& GRIT &55.8$\pm0.99$\\
\bottomrule
\end{tabular}

%% file: tables/diff_LLM.tex
\begin{tabular}{lcccccccccc}
\toprule
APM & SUM  & Bleu-1 & Bleu-2 & Bleu-3 & Bleu-4 & ROUGE-L & METEOR & CIDEr & SPICE&Execution Rate\\
\midrule
\multirow{3}{*}{BERT}
&OFA & \textbf{57.8}$\pm0.92$ & \textbf{43.8}$\pm1.02$ & \textbf{35.4}$\pm0.63$ & \textbf{23.5}$\pm0.77$ & 60.1$\pm0.41$ & \textbf{34.8}$\pm0.62$ & \textbf{31.8}$\pm1.31$ & \textbf{46.8}$\pm0.80$ & \textbf{77.8}$\pm3.26$ \\
&BLIP & 51.3$\pm0.31$ & 42.4$\pm0.54$ & 32.3$\pm0.66$ &22.3$\pm0.31$& 60.7$\pm0.63$ & 34.4$\pm0.75$ & 31.2$\pm0.87$ & 43.2$\pm0.97$&76.3$\pm5.22$\\
&GRIT & 52.9$\pm0.18$ & 41.6$\pm0.87$ & 32.4$\pm0.72$ & 22.1$\pm0.68$ & \textbf{61.6}$\pm0.53$ & 32.1$\pm0.33$ & 31.1$\pm0.25$ & 43.1$\pm0.76$ & 73.1$\pm3.11$\\
\midrule
\multirow{3}{*}{RoBERTa}
& OFA & \textbf{57.7}$\pm0.01$ & \textbf{43.2}$\pm0.00$ & \textbf{35.6}$\pm0.48$ & \textbf{24.1}$\pm0.36$ & 59.9$\pm0.26$ & \textbf{34.7}$\pm0.51$ & 31.4$\pm0.47$ & \textbf{47.3}$\pm0.38$ & 75.4$\pm3.86$ \\
& BLIP & 50.5$\pm0.71$ & 41.1$\pm0.29$ & 32.0$\pm0.11$ & 23.5$\pm0.64$ & \textbf{61.1}$\pm0.88$ & 33.0$\pm0.70$ & \textbf{31.8}$\pm0.81$ & 42.9$\pm0.94$ & \textbf{77.7}$\pm0.71$\\
& GRIT & 53.1$\pm1.02$ & 42.0$\pm0.90$ & 34.1$\pm1.01$ & 23.1$\pm1.22$ & 60.4$\pm1.92$ & 31.5$\pm0.59$ & 31.5$\pm1.42$ & 42.8$\pm1.77$ & 75.4$\pm4.39$\\
\midrule
\multirow{3}{*}{BART}
& OFA & \textbf{59.5}$\pm0.09$ & \textbf{45.9}$\pm0.31$ & \textbf{39.8}$\pm0.37$ & \textbf{28.1}$\pm0.72$ & 61.3$\pm0.65$ & \textbf{37.2}$\pm0.69$ & \textbf{34.4}$\pm0.78$ & \textbf{47.0}$\pm0.88$ & \textbf{79.0}$\pm1.91$\\
& BLIP & 52.9$\pm0.80$ & 44.3$\pm0.52$ & 35.5$\pm0.49$ & 25.3$\pm0.62$ & 62.2$\pm1.12$ & 35.3$\pm1.62$ & 32.0$\pm0.97$ & 44.5$\pm0.88$ & 76.0$\pm1.98$ \\
& GRIT & 54.2$\pm1.68$ & 43.2$\pm1.85$ & 33.6$\pm1.60$ & 25.3$\pm0.93$ & \textbf{62.7}$\pm1.85$ & 33.8$\pm0.62$ & 33.7$\pm0.74$ & 44.7$\pm1.12$ & 77.9$\pm1.77$\\
\bottomrule
\end{tabular}

%% file: tables/reinforce_sum.tex
\begin{tabular}{lccccccccc}
\toprule
SUM  & Env-1 & Env-2 & Env-3 & Env-4 & Env-5 & Env-6 & Env-7 & Average \\
\midrule
OFA & 50.1$\pm0.65$ &50.3$\pm0.52$ &51.5$\pm0.48$ &\textbf{57.8}$\pm0.88$ &55.2$\pm0.00$ &56.6$\pm0.37$ &\textbf{59.3}$\pm0.48$ &54.4$\pm0.55$       \\
BLIP &\textbf{52.7}$\pm0.78$ &\textbf{53.4}$\pm1.00$ &\textbf{53.5}$\pm0.92$ &55.6$\pm0.68$ &\textbf{60.1}$\pm0.49$ &\textbf{59.3}$\pm0.91$ &49.9$\pm0.90$ &\textbf{54.9}$\pm1.99$ \\
GRIT &38.7$\pm1.02$ &40.0$\pm1.11$ &51.3$\pm0.99$ &48.2$\pm0.90$ &46.5$\pm0.85$ &55.8$\pm0.70$ &45.3$\pm1.08$ &46.5$\pm2.01$ \\
\bottomrule
\end{tabular}

%% file: appendix.tex
\section{Algorithms of Fine-tuning SUM and APM with Imitation Learning or REINFORCE}\label{sec:algorithm_code}

\vspace{-5pt}
We provide the pseudo code for training SUM and APM in this section.
\vspace{-5pt}

\begin{minipage}{0.44\textwidth}
\begin{algorithm}[H]
    \centering
    \caption{Fine-tuning SUM}\label{alg:SUM}
    \footnotesize
\begin{algorithmic}
\STATE Initialize pretrained SUM model
\STATE Load VirtualHome dataset for fine-tuning
\FOR{$n$ in $\text{num}\_\text{epochs}$}
\FOR{$\text{Image}_t$ and $\text{Caption}_t$ in $\text{batch}_n$}
\STATE 1. $\hat{\text{Caption}_t} = \text{SUM}(\text{Image}_t)$
\STATE 2. $\text{Loss}_{{XE}_t}(\theta_t) = L_{XE}(\text{Caption}_t, \hat{\text{Caption}_t})$
\STATE 3. $\theta_t \leftarrow \theta_t - \alpha \nabla_{\theta_t} L(\text{Caption}_t,\hat{\text{Caption}_t})$
\ENDFOR
\REPEAT
\STATE{Steps 1 through 3}
\UNTIL{$\text{max}(\text{num}\_\text{epochs})$ or convergence}
\ENDFOR
\end{algorithmic}
\end{algorithm}
\end{minipage}
\hfill
\begin{minipage}{0.54\textwidth}
\begin{algorithm}[H]
    \centering
    \caption{Fine-tuning APM with Imitation Learning}\label{alg:APM_sup}
    \footnotesize
\begin{algorithmic}
\STATE Initialize fine-tuned SUM and pretrained APM
\STATE Load VirtualHome dataset for fine-tuning
\FOR{$n$ in $\text{num}\_\text{epochs}$}
\FOR{$\text{Image}_t$, $\text{Caption}_t$ $\text{Action}_t$ in $\text{batch}_n$}
\STATE 1. $\hat{\text{Caption}_t} = \text{SUM}(\text{Image}_t)$
\STATE 2. $\hat{\text{Action}_{t+1}} = \text{APM}(\hat{\text{Caption}_t}, \text{Action}_t)$
\STATE 3. $\text{Loss}_{{XE}_t}(\theta_t) = L_{XE}(\text{Action}_t, \hat{\text{Action}_{t+1}})$
\STATE 4. $\theta_t \leftarrow \theta_t - \alpha \nabla_{\theta_t} L_{XE}(\text{Action}_t,\hat{\text{Action}_{t+1}})$
\ENDFOR
\REPEAT
\STATE{Steps 1 through 3}
\UNTIL{$\text{max}(\text{num}\_\text{epochs})$ or convergence}
\ENDFOR
\end{algorithmic}
\end{algorithm}
\end{minipage}

\begin{algorithm}
\caption{Fine-tuning APM with REINFORCE}\label{alg:APM_reinforce}
\begin{algorithmic}
\STATE Initialize fine-tuned SUM, pretrained APM, and VirtualHome environment (env)
\STATE Load VirtualHome dataset for fine-tuning
\FOR{$n$ in $\text{num}\_\text{epochs}$}
\STATE $\text{Trajectories}_t = [~]$
\STATE $\text{state} = env.reset()$
\FOR{$\text{Image}_t$, $\text{Caption}_t$ $\text{Action}_t$ in $\text{batch}_n$}
\STATE 1. $\hat{\text{Caption}_t} = \text{SUM}(\text{Image}_t)$
\STATE 2. $\hat{\text{Action}_{t}} = \text{APM}(\hat{\text{Caption}_t}, \text{Action}_t)$
\STATE 3. $\text{Trajectories}_t.append(\hat{\text{Action}_{t}})$
\ENDFOR
\STATE $sort(\text{Trajectories}_t) \text{ by Task ID}$ 
\FOR{$i$ in $\text{range(len}(\text{Trajectories}_t))$}
\STATE 4. $\hat{\text{Action}_t} = \text{sample\_action}(\text{Trajectories}_t[i])$
\STATE 5. $\text{Reward}_t = env.step(\text{Action}_t, \hat{\text{Action}_t})$
\STATE 6. Compute $\nabla_{\theta_t} \log P(\hat{\text{Action}_t}|\text{Action}_t)$
\STATE 7. $\theta_t \leftarrow \theta_t + \alpha r \nabla_{\theta_t} \log P(\hat{\text{Action}_t}|\text{Action}_t)$
\ENDFOR
\REPEAT
\STATE{Steps 1 through 7}
\UNTIL{$\text{max}(\text{num}\_\text{epochs})$ or convergence}
\ENDFOR
\end{algorithmic}
\end{algorithm}

\section{Experimental Setup}\label{sec:appendix-experimental-setup}

\paragraph{SUM Setting}

For SUM, we use the following image captioning models to serve as SUM: OFA \citep{wang_2022_unifying}, BLIP \citep{li_2022_blip}, and GRIT \citep{Nguyen2022GRITFA}.
Both OFA and BLIP are pretrained on the same five datasets, while the GRIT model \citep{Nguyen2022GRITFA} is pretrained on a different combination of datasets.
For OFA, we adopted OFA$_{Large}$ due to its superior performance in five variations. OFA$_{Large}$ wields ResNet152 \citep{he_2015_deep} modules with 472M parameters and 12 encoders and decoder layers.
For BLIP, we used ViT-L/16 as the image encoder due to its better performance. 
For GRIP, we follow \citet{Nguyen2022GRITFA} which utilizes the Deformable DETR \citep{DBLP:journals/corr/abs-2010-04159} framework. 
Note that in our study we want SUM to generate captions that not only describe the scene but also try to derive action from it. We observe that adding the prompt "a picture of " following \citet{wang_2021_simvlm} causes the model to be biased in solely describing the scene, which would in turn not be helpful for generating actionable captions. 
Therefore, we remove prompts in the SUM setting.
We load pretrained models and fine-tune them for 7 epochs on our collected VirtualHome dataset. We keep the hyper-parameters consistent with the original implementations \citep{li_2022_blip, wang_2022_unifying,Nguyen2022GRITFA}. 

\vspace{-0.1in}

\paragraph{APM Setting}

We take LLM to act as the sole component in our APM.
The goal of APM is to generate executable programs for the VirtualHome simulator.
We deem the program outputted by the APM executable if the agent in the VirtualHome simulator is able to understand and perform the action.
When the action is executed by the agent, the simulator is then directed to output images and captions that are synonymous with the input of SUM. 
The output hidden layers of SUM acts as the input embeddings to the APM, while the tokenized executable actions act as labels.
The last hidden layer of APM acts as input embeddings for the tokenizer and generates token identifiers. The token identifiers are finally decoded into programmable actions that are fed into the VirtualHome simulator.

\section{Experiment Parameters}
\vspace{-10pt}
\begin{table*}[htp]\small
\centering
\caption{Experiment parameters used in SUMs, where the best ones are marked in bold}
\vspace{5pt}
\begin{adjustbox}{width=0.99\linewidth}
\begin{tabular}{lccccccc}
\toprule
SUM & Batch Size&Encoder Layers&Att. Heads&Learning Rate&Dropout&Epochs\\
\midrule
OFA
& [4, \textbf{8}, 16, 32]
& [\textbf{24}]
& [\textbf{16}]
&[1e-4, \textbf{1e-5}, 1e-7]
&[\textbf{0.1}, 0.2, 0.3]
&[2, 5, \textbf{10}, 20, 50]
 \\
BLIP
& [8, 16, \textbf{32}, 64]
& [\textbf{12}]
& [\textbf{12}]
&[1e-4, \textbf{1e-5}, 1e-7]
&[0.1, 0.2, \textbf{0.3}]
&[2, \textbf{5}, 10, 20, 50]
 \\
 GRIT
& [4, 8, 16, \textbf{32}]
& [\textbf{6}]
& [\textbf{8}]
&[\textbf{1e-4}, 1e-5, 1e-6]
&[0.1, \textbf{0.2}, 0.3]
&[2, 5, \textbf{10}, 20, 50]
 \\
\bottomrule
\end{tabular}
\end{adjustbox}
\label{Table:exp_param_SUM}
\vspace{-10pt}
\end{table*}
\vspace{-10pt}
\begin{table*}[htp]\small
\centering
\caption{Experiment parameters used in Supervised APMs, where the best ones are marked in bold}
\vspace{5pt}
\begin{adjustbox}{width=0.99\linewidth}
\begin{tabular}{lccccccc}
\toprule
APM & Batch Size&Encoder Layers&Att. Heads&Learning Rate&Dropout&Epochs\\ 
\midrule
BERT
& [4, \textbf{8}, 16, 32]
& [\textbf{12}]
& [\textbf{12}]
&[1e-4, \textbf{1e-5}, 1e-7]
&[0.1, 0.2, \textbf{0.3}]
&[2, 5, \textbf{10}, 20, 50]
 \\
BART
& [8, 16, \textbf{32}, 64]
& [\textbf{12}]
& [\textbf{16}]
&[1e-4, \textbf{1e-5}, 1e-7]
&[0.1, 0.2, \textbf{0.3}]
&[2, 5, \textbf{10}, 20, 50]
 \\
RoBERTa
& [4, 8, 16, \textbf{32}]
& [\textbf{12}]
& [\textbf{12}]
&[\textbf{1e-4}, 1e-5, 1e-7]
&[0.1, 0.2, \textbf{0.3}]
&[2, 5, \textbf{10}, 20, 50]
 \\
\bottomrule
\end{tabular}
\end{adjustbox}
\label{Table:exp_param_APM_sup}
\vspace{-10pt}
\end{table*}
\vspace{-10pt}
\begin{table*}[htp]\small
\centering
\caption{Experiment parameters used in REINFORCE APMs, where the best ones are marked in bold}
\vspace{5pt}
\begin{adjustbox}{width=0.99\linewidth}
\begin{tabular}{lccccccc}
\toprule
APM & Batch Size&Encoder Layers&Att. Heads&Learning Rate&Dropout&Epochs\\ 
\midrule
BERT
& [4, \textbf{8}, 16, 32]
& [\textbf{12}]
& [\textbf{12}]
&[1e-4, \textbf{1e-5}, 1e-7]
&[0.1, 0.2, \textbf{0.3}]
&[2, 5, \textbf{10}, 20, 50]
 \\
BART
& [8, 16, \textbf{32}, 64]
& [\textbf{12}]
& [\textbf{16}]
&[1e-4, \textbf{1e-5}, 1e-7]
&[0.1, 0.2, \textbf{0.3}]
&[2, 5, \textbf{10}, 20, 50]
 \\
RoBERTa
& [4, 8, 16, \textbf{32}]
& [\textbf{12}]
& [\textbf{12}]
&[1e-4, \textbf{1e-5}, 1e-7]
&[0.1, \textbf{0.2}, 0.3]
&[2, 5, \textbf{10}, 20, 50]
 \\
\bottomrule
\end{tabular}
\end{adjustbox}
\label{Table:exp_param_APM_rein}
\vspace{-10pt}
\end{table*}
\vspace{5pt}

\section{More Experimental Results}

\paragraph{Fine-tuning performance on in-distribution tasks and unseen tasks}

To further support our findings, we conducted additional experiments that tested the fine-tuning performance on in-distribution tasks and unseen tasks in the VirtualHome environment following the setting in \citet{Li2022pretrainedLM}. \citet{Li2022pretrainedLM} used reinforcement learning to adapt to downstream tasks. It’s important to note that \citet{Li2022pretrainedLM} used oracle text-based inputs that summarize the current observation, whereas we use raw image inputs and understand the scene with our fine-tuned SUM module. We measure the performance with the episode success rate and summarize the main comparison results with \citet{Li2022pretrainedLM}) in Table~\ref{table:in-distribution-li}. Our results show that when fine-tuning with REINFORCE, our method outperforms \citet{Li2022pretrainedLM} in both in-distribution tasks and novel tasks. Additionally, when expert data is available in the downstream tasks, fine-tuning with imitation learning outperforms the REINFORCE approach.

\begin{table}[htp]
\centering
\caption{Comparison of episode success rate.}
\vspace{-5pt}
\begin{adjustbox}{width=0.6\linewidth}
\begin{tabular}{lcc}
\toprule
Method& In-Distribution Tasks &Novel Tasks \\
\midrule
\citet{Li2022pretrainedLM} & 53.7 &27.8 \\
Ours (REINFORCE)     &58.4    & 33.7 \\
Ours (Imitation Learning)     & 68.4    &  44.8 \\
\bottomrule
\end{tabular}
\end{adjustbox}
\label{table:in-distribution-li}
\vspace{-5pt}
\end{table}

\begin{table}[htp]
\centering
\caption{ Our fine-tuning results for different SUM/APM configurations in in-distribution and novel tasks, as well as using REINFORCE and imitation learning strategies. We measure the performance based on the episode success rate.}
\begin{adjustbox}{width=0.99\linewidth}
\begin{tabular}{lccccc}
\toprule
 SUM  &  APM   & In-Distribution REINFORCE  & Novel Tasks REINFORCE   &  In-Distribution Imitation  &  Novel Tasks Imitation \\
\midrule
\multirow{3}{*}{OFA}
&BERT & 56.1 & 31.4 & 65.2 & 40.7 \\
& BART & \textbf{58.4} & \textbf{33.7} & \textbf{68.4} & \textbf{44.8} \\
 & RoBERTa &  51.7 & 32.3 &  66.0 & 42.8 \\
\midrule
\multirow{3}{*}{BLIP}
& BERT & 53.7  & 28.5 & 61.1  & 39.5 \\
& BART & 55.2 & 31.2 & 64.3 & 40.3 \\
& RoBERTa & 50.6 & 29.3 & 62.8 & 39.8 \\
\midrule
\multirow{3}{*}{GRIT}
& BERT & 50.5  & 28.8 & 61.3  & 40.4 \\
& BART & 51.2  & 30.0 & 63.7  & 39.6 \\
& RoBERTa & 49.0  & 27.1 & 59.2  & 38.7 \\
\bottomrule
\end{tabular}
\end{adjustbox}
\label{table:in-distribution}
\vspace{-10pt}
\end{table}

\paragraph{Importance and necessity of fine-tuning}

To underscore the importance and necessity of fine-tuning, we present additional zero-shot testing performances without fine-tuning in Table~\ref{table:ft-execution} and Table~\ref{table:ft-success}. Our findings reveal that the episode success rate and action execution rates are significantly lower without fine-tuning in both methods, which highlights the crucial role that fine-tuning plays in improving performance.

\begin{table}[H]
\centering
\caption{Comparison action execution rates in zero-shot and fine-tuned settings using both REINFORCE and Imitation Learning.}
\begin{adjustbox}{width=0.65\linewidth}
\begin{tabular}{lcccc}
\toprule
Method& APM &SUM & REINFORCE & Imitation Learning \\
\midrule
1    & Zero-shot  & Zero-shot  &    0.1    &        0.1 \\
2    & Zero-shot  & Fine-tuned &   14.5    &        21.4  \\
3    & Fine-tuned & Zero-shot  &    5.8    &        6.9  \\
4    & Fine-tuned & Fine-tuned &   57.2    &        77.8  \\
\bottomrule
\end{tabular}
\end{adjustbox}
\label{table:ft-execution}
\vspace{-10pt}
\end{table}

\begin{table}[H]
\centering
\caption{Comparison episode success rate in zero-shot and fine-tuned settings using both REINFORCE and Imitation Learning.}
\begin{adjustbox}{width=0.65\linewidth}
\begin{tabular}{lcccc}
\toprule
Method& APM &SUM & REINFORCE & Imitation Learning \\
\midrule
1    & Zero-shot  & Zero-shot  &    0.7    &        0.7   \\
2    & Zero-shot  & Fine-tuned &   16.7    &        19.5  \\
3    & Fine-tuned & Zero-shot  &    7.7    &        8.7    \\
4    & Fine-tuned & Fine-tuned &   58.4    &        76.8  \\
\bottomrule
\end{tabular}
\end{adjustbox}
\label{table:ft-success}
\vspace{-10pt}
\end{table}

\section{More Related Work}\label{sec:appendix-related-work}

\paragraph{Multimodal Learning}

Formalized multimodal learning research dates back to 1989 when \cite{yuhas_1989_integration} conducted an experiment that built off the McGurk Effect for audio-visual speech recognition using neural networks \citep{tiippana_2014_what,McGurk1976HearingLA}. Researchers in NLP and CV collaborated to make large and multimodal datasets available, catering to specific downstream tasks, such as classification, translation, and detection. 
In correlation, improvements in LLMs opened the gates to include other modalities of data, most frequently visual data \citep{wang_2022_unifying, Nguyen2022GRITFA, li_2022_blip, wang_2021_simvlm, Shah2022LMNavRN, Zhang_2021_CVPR, DBLP:journals/corr/abs-2012-06946}.
By utilizing the learned embeddings that have been pretrained on both language and image datasets, vision-language models are able to perform very well.
Within the above success, image captioning has been an important task in multimodal learning, which aims at generating textual descriptions for the given images. 

\paragraph{Visual Feedback in Robot Learning}
Visual feedback is commonly used in robot learning. 
\citet{Gothoskar2020LearningAG} learned a generative model from actions to image observations of features to control a robot from visual feedback. 
\citet{Ma2022VIPTU} proposed a self-supervised pretrained visual representation model which is capable of generating dense and smooth reward functions for unseen robotic tasks. 
\citet{Strokina2022VisualRF} reviewed the methods of reward estimation and visual representations used in learning-based approaches for robotics applications. 
\citet{Mohtasib2021ASO} studied the performance of dense, sparse, visually dense, and visually sparse rewards in deep RL.

%% file: main.bbl
\begin{thebibliography}{77}
\providecommand{\natexlab}[1]{#1}
\providecommand{\url}[1]{\texttt{#1}}
\expandafter\ifx\csname urlstyle\endcsname\relax
  \providecommand{\doi}[1]{doi: #1}\else
  \providecommand{\doi}{doi: \begingroup \urlstyle{rm}\Url}\fi

\bibitem[Vaswani et~al.(2017)Vaswani, Shazeer, Parmar, Uszkoreit, Jones, Gomez,
  Kaiser, and Polosukhin]{Vaswani2017AttentionIA}
A.~Vaswani, N.~M. Shazeer, N.~Parmar, J.~Uszkoreit, L.~Jones, A.~N. Gomez,
  L.~Kaiser, and I.~Polosukhin.
\newblock Attention is all you need.
\newblock \emph{ArXiv}, abs/1706.03762, 2017.

\bibitem[Devlin et~al.(2019)Devlin, Chang, Lee, and
  Toutanova]{Devlin2019BERTPO}
J.~Devlin, M.-W. Chang, K.~Lee, and K.~Toutanova.
\newblock Bert: Pre-training of deep bidirectional transformers for language
  understanding.
\newblock In \emph{NAACL}, 2019.

\bibitem[Liu et~al.(2019)Liu, Ott, Goyal, Du, Joshi, Chen, Levy, Lewis,
  Zettlemoyer, and Stoyanov]{Liu2019RoBERTaAR}
Y.~Liu, M.~Ott, N.~Goyal, J.~Du, M.~Joshi, D.~Chen, O.~Levy, M.~Lewis,
  L.~Zettlemoyer, and V.~Stoyanov.
\newblock Roberta: A robustly optimized bert pretraining approach.
\newblock \emph{ArXiv}, abs/1907.11692, 2019.

\bibitem[Brown et~al.(2020)]{Brown2020LanguageMA}
T.~B. Brown et~al.
\newblock Language models are few-shot learners.
\newblock \emph{ArXiv}, abs/2005.14165, 2020.

\bibitem[Liang et~al.(2022)Liang, Huang, Xia, Xu, Hausman, Ichter, Florence,
  and Zeng]{Liang2022CodeAP}
J.~Liang, W.~Huang, F.~Xia, P.~Xu, K.~Hausman, B.~Ichter, P.~R. Florence, and
  A.~Zeng.
\newblock Code as policies: Language model programs for embodied control.
\newblock \emph{ArXiv}, abs/2209.07753, 2022.

\bibitem[Ahn et~al.(2022)]{Ahn2022DoAI}
M.~Ahn et~al.
\newblock Do as i can, not as i say: Grounding language in robotic affordances.
\newblock \emph{ArXiv}, abs/2204.01691, 2022.

\bibitem[Zeng et~al.(2022)Zeng, Wong, Welker, Choromanski, Tombari, Purohit,
  Ryoo, Sindhwani, Lee, Vanhoucke, and Florence]{Zeng2022SocraticMC}
A.~Zeng, A.~S. Wong, S.~Welker, K.~Choromanski, F.~Tombari, A.~Purohit, M.~S.
  Ryoo, V.~Sindhwani, J.~Lee, V.~Vanhoucke, and P.~R. Florence.
\newblock Socratic models: Composing zero-shot multimodal reasoning with
  language.
\newblock \emph{ArXiv}, abs/2204.00598, 2022.

\bibitem[Zellers et~al.(2021)Zellers, Holtzman, Peters, Mottaghi, Kembhavi,
  Farhadi, and Choi]{Zellers2021PIGLeTLG}
R.~Zellers, A.~Holtzman, M.~E. Peters, R.~Mottaghi, A.~Kembhavi, A.~Farhadi,
  and Y.~Choi.
\newblock Piglet: Language grounding through neuro-symbolic interaction in a 3d
  world.
\newblock In \emph{ACL}, 2021.

\bibitem[Li et~al.(2022)Li, Puig, Du, Wang, Aky{\"u}rek, Torralba, Andreas, and
  Mordatch]{Li2022pretrainedLM}
S.~Li, X.~Puig, Y.~Du, C.~J. Wang, E.~Aky{\"u}rek, A.~Torralba, J.~Andreas, and
  I.~Mordatch.
\newblock Pre-trained language models for interactive decision-making.
\newblock \emph{ArXiv}, abs/2202.01771, 2022.

\bibitem[Huang et~al.(2022)Huang, Abbeel, Pathak, and
  Mordatch]{Huang2022LanguageMA}
W.~Huang, P.~Abbeel, D.~Pathak, and I.~Mordatch.
\newblock Language models as zero-shot planners: Extracting actionable
  knowledge for embodied agents.
\newblock In \emph{ICML}, 2022.

\bibitem[Fried et~al.(2022)Fried, Aghajanyan, Lin, Wang, Wallace, Shi, Zhong,
  tau Yih, Zettlemoyer, and Lewis]{Fried2022InCoderAG}
D.~Fried, A.~Aghajanyan, J.~Lin, S.~I. Wang, E.~Wallace, F.~Shi, R.~Zhong,
  W.~tau Yih, L.~Zettlemoyer, and M.~Lewis.
\newblock Incoder: A generative model for code infilling and synthesis.
\newblock \emph{ArXiv}, abs/2204.05999, 2022.

\bibitem[Kaplan et~al.(2020)Kaplan, McCandlish, Henighan, Brown, Chess, Child,
  Gray, Radford, Wu, and Amodei]{Kaplan2020ScalingLF}
J.~Kaplan, S.~McCandlish, T.~J. Henighan, T.~B. Brown, B.~Chess, R.~Child,
  S.~Gray, A.~Radford, J.~Wu, and D.~Amodei.
\newblock Scaling laws for neural language models.
\newblock \emph{ArXiv}, abs/2001.08361, 2020.

\bibitem[Radford et~al.(2021)Radford, Kim, Hallacy, Ramesh, Goh, Agarwal,
  Sastry, Askell, Mishkin, Clark, Krueger, and
  Sutskever]{Radford2021LearningTV}
A.~Radford, J.~W. Kim, C.~Hallacy, A.~Ramesh, G.~Goh, S.~Agarwal, G.~Sastry,
  A.~Askell, P.~Mishkin, J.~Clark, G.~Krueger, and I.~Sutskever.
\newblock Learning transferable visual models from natural language
  supervision.
\newblock In \emph{ICML}, 2021.

\bibitem[Shah et~al.(2022)Shah, Osinski, Ichter, and Levine]{Shah2022LMNavRN}
D.~Shah, B.~Osinski, B.~Ichter, and S.~Levine.
\newblock Lm-nav: Robotic navigation with large pre-trained models of language,
  vision, and action.
\newblock \emph{ArXiv}, abs/2207.04429, 2022.

\bibitem[Huang et~al.(2022)]{Huang2022InnerME}
W.~Huang et~al.
\newblock Inner monologue: Embodied reasoning through planning with language
  models.
\newblock \emph{ArXiv}, abs/2207.05608, 2022.

\bibitem[Kant et~al.(2022)]{Kant2022HousekeepTV}
Y.~Kant et~al.
\newblock Housekeep: Tidying virtual households using commonsense reasoning.
\newblock \emph{ArXiv}, abs/2205.10712, 2022.

\bibitem[Shafiullah et~al.(2022)Shafiullah, Paxton, Pinto, Chintala, and
  Szlam]{Shafiullah2022CLIPFieldsWS}
N.~M.~M. Shafiullah, C.~Paxton, L.~Pinto, S.~Chintala, and A.~D. Szlam.
\newblock Clip-fields: Weakly supervised semantic fields for robotic memory.
\newblock \emph{ArXiv}, abs/2210.05663, 2022.

\bibitem[Jiang et~al.(2022)Jiang, Gupta, Zhang, Wang, Dou, Chen, Fei-Fei,
  Anandkumar, Zhu, and Fan]{Jiang2022VIMAGR}
Y.~Jiang, A.~Gupta, Z.~V. Zhang, G.~Wang, Y.~Dou, Y.~Chen, L.~Fei-Fei,
  A.~Anandkumar, Y.~Zhu, and L.~J. Fan.
\newblock Vima: General robot manipulation with multimodal prompts.
\newblock \emph{ArXiv}, abs/2210.03094, 2022.

\bibitem[Shridhar et~al.(2022)Shridhar, Manuelli, and
  Fox]{Shridhar2022PerceiverActorAM}
M.~Shridhar, L.~Manuelli, and D.~Fox.
\newblock Perceiver-actor: A multi-task transformer for robotic manipulation.
\newblock \emph{ArXiv}, abs/2209.05451, 2022.

\bibitem[Bucker et~al.(2022)Bucker, Figueredo, Haddadin, Kapoor, Ma, Vemprala,
  and Bonatti]{Bucker2022LaTTeLT}
A.~F.~C. Bucker, L.~F.~C. Figueredo, S.~Haddadin, A.~Kapoor, S.~Ma,
  S.~Vemprala, and R.~Bonatti.
\newblock Latte: Language trajectory transformer.
\newblock \emph{ArXiv}, abs/2208.02918, 2022.

\bibitem[Ren et~al.(2022)Ren, Govil, Yang, Narasimhan, and
  Majumdar]{Ren2022LeveragingLF}
A.~Z. Ren, B.~Govil, T.-Y. Yang, K.~Narasimhan, and A.~Majumdar.
\newblock Leveraging language for accelerated learning of tool manipulation.
\newblock \emph{ArXiv}, abs/2206.13074, 2022.

\bibitem[Tam et~al.(2022)Tam, Rabinowitz, Lampinen, Roy, Chan, Strouse, Wang,
  Banino, and Hill]{Tam2022SemanticEF}
A.~C. Tam, N.~C. Rabinowitz, A.~K. Lampinen, N.~A. Roy, S.~C.~Y. Chan,
  D.~Strouse, J.~X. Wang, A.~Banino, and F.~Hill.
\newblock Semantic exploration from language abstractions and pretrained
  representations.
\newblock \emph{ArXiv}, abs/2204.05080, 2022.

\bibitem[Khandelwal et~al.(2022)Khandelwal, Weihs, Mottaghi, and
  Kembhavi]{Khandelwal2022SimpleBE}
A.~Khandelwal, L.~Weihs, R.~Mottaghi, and A.~Kembhavi.
\newblock Simple but effective: Clip embeddings for embodied ai.
\newblock \emph{CVPR}, pages 14809--14818, 2022.

\bibitem[Shridhar et~al.(2021)Shridhar, Manuelli, and
  Fox]{Shridhar2021CLIPortWA}
M.~Shridhar, L.~Manuelli, and D.~Fox.
\newblock Cliport: What and where pathways for robotic manipulation.
\newblock \emph{ArXiv}, abs/2109.12098, 2021.

\bibitem[Lin et~al.(2022)Lin, Zhu, Chen, Liang, zhuo Liu, and
  Liang]{Lin2022ADAPTVN}
B.~Lin, Y.~Zhu, Z.~Chen, X.~Liang, J.~zhuo Liu, and X.~Liang.
\newblock Adapt: Vision-language navigation with modality-aligned action
  prompts.
\newblock \emph{CVPR}, pages 15375--15385, 2022.

\bibitem[Parisi et~al.(2022)Parisi, Rajeswaran, Purushwalkam, and
  Gupta]{Parisi2022TheUE}
S.~Parisi, A.~Rajeswaran, S.~Purushwalkam, and A.~K. Gupta.
\newblock The unsurprising effectiveness of pre-trained vision models for
  control.
\newblock In \emph{ICML}, 2022.

\bibitem[Gadre et~al.(2022)Gadre, Wortsman, Ilharco, Schmidt, and
  Song]{Gadre2022CLIPOW}
S.~Y. Gadre, M.~Wortsman, G.~Ilharco, L.~Schmidt, and S.~Song.
\newblock Clip on wheels: Zero-shot object navigation as object localization
  and exploration.
\newblock \emph{ArXiv}, abs/2203.10421, 2022.

\bibitem[Hong et~al.(2021)Hong, Wu, Qi, Rodriguez-Opazo, and
  Gould]{Hong2021VLNBERTAR}
Y.~Hong, Q.~Wu, Y.~Qi, C.~Rodriguez-Opazo, and S.~Gould.
\newblock Vln-bert: A recurrent vision-and-language bert for navigation.
\newblock \emph{CVPR}, pages 1643--1653, 2021.

\bibitem[Majumdar et~al.(2020)Majumdar, Shrivastava, Lee, Anderson, Parikh, and
  Batra]{Majumdar2020ImprovingVN}
A.~Majumdar, A.~Shrivastava, S.~Lee, P.~Anderson, D.~Parikh, and D.~Batra.
\newblock Improving vision-and-language navigation with image-text pairs from
  the web.
\newblock \emph{ArXiv}, abs/2004.14973, 2020.

\bibitem[Singh et~al.(2022)Singh, Blukis, Mousavian, Goyal, Xu, Tremblay, Fox,
  Thomason, and Garg]{singh2022progprompt}
I.~Singh, V.~Blukis, A.~Mousavian, A.~Goyal, D.~Xu, J.~Tremblay, D.~Fox,
  J.~Thomason, and A.~Garg.
\newblock Progprompt: Generating situated robot task plans using large language
  models.
\newblock \emph{arXiv preprint arXiv:2209.11302}, 2022.

\bibitem[Xiao et~al.(2022)Xiao, Chan, Sermanet, Wahid, Brohan, Hausman, Levine,
  and Tompson]{xiao2022robotic}
T.~Xiao, H.~Chan, P.~Sermanet, A.~Wahid, A.~Brohan, K.~Hausman, S.~Levine, and
  J.~Tompson.
\newblock Robotic skill acquisition via instruction augmentation with
  vision-language models.
\newblock \emph{arXiv preprint arXiv:2211.11736}, 2022.

\bibitem[Fan et~al.(2022)Fan, Wang, Jiang, Mandlekar, Yang, Zhu, Tang, Huang,
  Zhu, and Anandkumar]{fan2022minedojo}
L.~Fan, G.~Wang, Y.~Jiang, A.~Mandlekar, Y.~Yang, H.~Zhu, A.~Tang, D.-A. Huang,
  Y.~Zhu, and A.~Anandkumar.
\newblock Minedojo: Building open-ended embodied agents with internet-scale
  knowledge.
\newblock \emph{arXiv preprint arXiv:2206.08853}, 2022.

\bibitem[Howard and Ruder(2018)]{Howard2018UniversalLM}
J.~Howard and S.~Ruder.
\newblock Universal language model fine-tuning for text classification.
\newblock In \emph{ACL}, 2018.

\bibitem[Peters et~al.(2018)Peters, Neumann, Iyyer, Gardner, Clark, Lee, and
  Zettlemoyer]{Peters2018DeepCW}
M.~E. Peters, M.~Neumann, M.~Iyyer, M.~Gardner, C.~Clark, K.~Lee, and
  L.~Zettlemoyer.
\newblock Deep contextualized word representations.
\newblock In \emph{NAACL}, 2018.

\bibitem[Ruder(2021)]{SEBASTIANRecent}
S.~Ruder.
\newblock Recent advances in language model fine-tuning.
\newblock 2021.

\bibitem[Raffel et~al.(2019)Raffel, Shazeer, Roberts, Lee, Narang, Matena,
  Zhou, Li, and Liu]{Raffel2019ExploringTL}
C.~Raffel, N.~M. Shazeer, A.~Roberts, K.~Lee, S.~Narang, M.~Matena, Y.~Zhou,
  W.~Li, and P.~J. Liu.
\newblock Exploring the limits of transfer learning with a unified text-to-text
  transformer.
\newblock \emph{ArXiv}, abs/1910.10683, 2019.

\bibitem[Hendrycks et~al.(2020)Hendrycks, Liu, Wallace, Dziedzic, Krishnan, and
  Song]{Hendrycks2020pretrainedTI}
D.~Hendrycks, X.~Liu, E.~Wallace, A.~Dziedzic, R.~Krishnan, and D.~X. Song.
\newblock Pretrained transformers improve out-of-distribution robustness.
\newblock In \emph{ACL}, 2020.

\bibitem[Dodge et~al.(2020)Dodge, Ilharco, Schwartz, Farhadi, Hajishirzi, and
  Smith]{Dodge2020FineTuningPL}
J.~Dodge, G.~Ilharco, R.~Schwartz, A.~Farhadi, H.~Hajishirzi, and N.~A. Smith.
\newblock Fine-tuning pretrained language models: Weight initializations, data
  orders, and early stopping.
\newblock \emph{ArXiv}, abs/2002.06305, 2020.

\bibitem[Ke et~al.(2019)Ke, Pei, Li, Shen, and Tai]{Ke2019ReflectiveDN}
L.~Ke, W.~Pei, R.~Li, X.~Shen, and Y.-W. Tai.
\newblock Reflective decoding network for image captioning.
\newblock \emph{ICCV}, pages 8887--8896, 2019.

\bibitem[Stefanini et~al.(2021)Stefanini, Cornia, Baraldi, Cascianelli,
  Fiameni, and Cucchiara]{Stefanini2021FromST}
M.~Stefanini, M.~Cornia, L.~Baraldi, S.~Cascianelli, G.~Fiameni, and
  R.~Cucchiara.
\newblock From show to tell: A survey on deep learning-based image captioning.
\newblock \emph{IEEE Transactions on Pattern Analysis and Machine
  Intelligence}, 45:\penalty0 539--559, 2021.

\bibitem[Vaswani et~al.(2017)Vaswani, Shazeer, Parmar, Uszkoreit, Jones, Gomez,
  Kaiser, and Polosukhin]{vaswani_2017_attention}
A.~Vaswani, N.~Shazeer, N.~Parmar, J.~Uszkoreit, L.~Jones, A.~N. Gomez,
  L.~Kaiser, and I.~Polosukhin.
\newblock Attention is all you need, 2017.

\bibitem[Tan and Bansal(2019)]{Tan2019LXMERTLC}
H.~H. Tan and M.~Bansal.
\newblock Lxmert: Learning cross-modality encoder representations from
  transformers.
\newblock \emph{ArXiv}, abs/1908.07490, 2019.

\bibitem[Lu et~al.(2019)Lu, Batra, Parikh, and Lee]{Lu2019ViLBERTPT}
J.~Lu, D.~Batra, D.~Parikh, and S.~Lee.
\newblock Vilbert: Pretraining task-agnostic visiolinguistic representations
  for vision-and-language tasks.
\newblock In \emph{Neural Information Processing Systems}, 2019.

\bibitem[Li et~al.(2020)Li, Yin, Li, Hu, Zhang, Zhang, Wang, Hu, Dong, Wei,
  Choi, and Gao]{Li2020OscarOA}
X.~Li, X.~Yin, C.~Li, X.~Hu, P.~Zhang, L.~Zhang, L.~Wang, H.~Hu, L.~Dong,
  F.~Wei, Y.~Choi, and J.~Gao.
\newblock Oscar: Object-semantics aligned pre-training for vision-language
  tasks.
\newblock In \emph{ECCV}, 2020.

\bibitem[Zhou et~al.(2019)Zhou, Palangi, Zhang, Hu, Corso, and
  Gao]{Zhou2019UnifiedVP}
L.~Zhou, H.~Palangi, L.~Zhang, H.~Hu, J.~J. Corso, and J.~Gao.
\newblock Unified vision-language pre-training for image captioning and vqa.
\newblock \emph{ArXiv}, abs/1909.11059, 2019.

\bibitem[Lewis et~al.(2020)Lewis, Liu, Goyal, Ghazvininejad, Mohamed, Levy,
  Stoyanov, and Zettlemoyer]{Lewis2020BARTDS}
M.~Lewis, Y.~Liu, N.~Goyal, M.~Ghazvininejad, A.~Mohamed, O.~Levy, V.~Stoyanov,
  and L.~Zettlemoyer.
\newblock Bart: Denoising sequence-to-sequence pre-training for natural
  language generation, translation, and comprehension.
\newblock In \emph{ACL}, 2020.

\bibitem[Cho et~al.(2014)Cho, van Merrienboer, Çaglar G{\"u}lçehre, Bahdanau,
  Bougares, Schwenk, and Bengio]{Cho2014LearningPR}
K.~Cho, B.~van Merrienboer, Çaglar G{\"u}lçehre, D.~Bahdanau, F.~Bougares,
  H.~Schwenk, and Y.~Bengio.
\newblock Learning phrase representations using rnn encoder–decoder for
  statistical machine translation.
\newblock In \emph{EMNLP}, 2014.

\bibitem[Koehn(2007)]{Philipp2007}
P.~Koehn.
\newblock Statistical machine translation.
\newblock 2007.

\bibitem[Ranzato et~al.(2015)Ranzato, Chopra, Auli, and
  Zaremba]{Ranzato2015SequenceLT}
M.~Ranzato, S.~Chopra, M.~Auli, and W.~Zaremba.
\newblock Sequence level training with recurrent neural networks.
\newblock \emph{CoRR}, abs/1511.06732, 2015.

\bibitem[Williams(1992)]{Williams1992SimpleSG}
R.~J. Williams.
\newblock Simple statistical gradient-following algorithms for connectionist
  reinforcement learning.
\newblock \emph{Machine Learning}, 8:\penalty0 229--256, 1992.

\bibitem[Sutton et~al.(1999)Sutton, McAllester, Singh, and
  Mansour]{NIPS1999_464d828b}
R.~S. Sutton, D.~McAllester, S.~Singh, and Y.~Mansour.
\newblock Policy gradient methods for reinforcement learning with function
  approximation.
\newblock In S.~Solla, T.~Leen, and K.~M\"{u}ller, editors, \emph{NeurIPS},
  volume~12. MIT Press, 1999.

\bibitem[Rennie et~al.(2016)Rennie, Marcheret, Mroueh, Ross, and
  Goel]{Rennie2016SelfCriticalST}
S.~J. Rennie, E.~Marcheret, Y.~Mroueh, J.~Ross, and V.~Goel.
\newblock Self-critical sequence training for image captioning.
\newblock \emph{CVPR}, pages 1179--1195, 2016.

\bibitem[Cornia et~al.(2019)Cornia, Stefanini, Baraldi, and
  Cucchiara]{Cornia2019MeshedMemoryTF}
M.~Cornia, M.~Stefanini, L.~Baraldi, and R.~Cucchiara.
\newblock Meshed-memory transformer for image captioning.
\newblock \emph{CVPR}, pages 10575--10584, 2019.

\bibitem[Puig et~al.(2018{\natexlab{a}})Puig, Ra, Boben, Li, Wang, Fidler, and
  Torralba]{Puig2018VirtualHomeSH}
X.~Puig, K.~K. Ra, M.~Boben, J.~Li, T.~Wang, S.~Fidler, and A.~Torralba.
\newblock Virtualhome: Simulating household activities via programs.
\newblock \emph{CVPR}, pages 8494--8502, 2018{\natexlab{a}}.

\bibitem[Puig et~al.(2018{\natexlab{b}})Puig, Ra, Boben, Li, Wang, Fidler, and
  Torralba]{puig_2018_virtualhome}
X.~Puig, K.~Ra, M.~Boben, J.~Li, T.~Wang, S.~Fidler, and A.~Torralba.
\newblock Virtualhome: Simulating household activities via programs.
\newblock \emph{arXiv:1806.07011 [cs]}, 06 2018{\natexlab{b}}.

\bibitem[Liao et~al.(2019)Liao, Puig, Boben, Torralba, and
  Fidler]{liao_2019_synthesizing}
Y.-H. Liao, X.~Puig, M.~Boben, A.~Torralba, and S.~Fidler.
\newblock Synthesizing environment-aware activities via activity sketches, 06
  2019.

\bibitem[Papineni et~al.(2002)Papineni, Roukos, Ward, and
  Zhu]{Papineni2002BleuAM}
K.~Papineni, S.~Roukos, T.~Ward, and W.-J. Zhu.
\newblock Bleu: a method for automatic evaluation of machine translation.
\newblock In \emph{ACL}, 2002.

\bibitem[Lin(2004)]{Lin2004ROUGEAP}
C.-Y. Lin.
\newblock Rouge: A package for automatic evaluation of summaries.
\newblock In \emph{ACL 2004}, 2004.

\bibitem[Banerjee and Lavie(2005)]{Banerjee2005METEORAA}
S.~Banerjee and A.~Lavie.
\newblock Meteor: An automatic metric for mt evaluation with improved
  correlation with human judgments.
\newblock In \emph{IEEvaluation@ACL}, 2005.

\bibitem[Vedantam et~al.(2015)Vedantam, Zitnick, and
  Parikh]{vedantam_2015_cider}
R.~Vedantam, C.~L. Zitnick, and D.~Parikh.
\newblock Cider: Consensus-based image description evaluation.
\newblock \emph{arXiv:1411.5726 [cs]}, 06 2015.

\bibitem[Anderson et~al.(2016)Anderson, Fernando, Johnson, and
  Gould]{DBLP:journals/corr/AndersonFJG16}
P.~Anderson, B.~Fernando, M.~Johnson, and S.~Gould.
\newblock {SPICE:} semantic propositional image caption evaluation.
\newblock \emph{CoRR}, abs/1607.08822, 2016.

\bibitem[Song et~al.(2019)Song, Tan, Qin, Lu, and Liu]{song_2019_mass}
K.~Song, X.~Tan, T.~Qin, J.~Lu, and T.-Y. Liu.
\newblock Mass: Masked sequence to sequence pre-training for language
  generation.
\newblock \emph{arXiv:1905.02450 [cs]}, 06 2019.

\bibitem[Wang et~al.(2022)Wang, Yang, Men, Lin, Bai, Li, Ma, Zhou, Zhou, and
  Yang]{wang_2022_unifying}
P.~Wang, A.~Yang, R.~Men, J.~Lin, S.~Bai, Z.~Li, J.~Ma, C.~Zhou, J.~Zhou, and
  H.~Yang.
\newblock Unifying architectures, tasks, and modalities through a simple
  sequence-to-sequence learning framework.
\newblock \emph{arXiv:2202.03052 [cs]}, 02 2022.

\bibitem[Li et~al.(2022)Li, Li, Xiong, and Hoi]{li_2022_blip}
J.~Li, D.~Li, C.~Xiong, and S.~Hoi.
\newblock Blip: Bootstrapping language-image pre-training for unified
  vision-language understanding and generation.
\newblock \emph{arXiv:2201.12086 [cs]}, 02 2022.

\bibitem[Nguyen et~al.(2022)Nguyen, Suganuma, and Okatani]{Nguyen2022GRITFA}
V.-Q. Nguyen, M.~Suganuma, and T.~Okatani.
\newblock Grit: Faster and better image captioning transformer using dual
  visual features.
\newblock \emph{ArXiv}, abs/2207.09666, 2022.

\bibitem[He et~al.(2015)He, Zhang, Ren, and Sun]{he_2015_deep}
K.~He, X.~Zhang, S.~Ren, and J.~Sun.
\newblock Deep residual learning for image recognition, 12 2015.

\bibitem[Zhu et~al.(2020)Zhu, Su, Lu, Li, Wang, and
  Dai]{DBLP:journals/corr/abs-2010-04159}
X.~Zhu, W.~Su, L.~Lu, B.~Li, X.~Wang, and J.~Dai.
\newblock Deformable {DETR:} deformable transformers for end-to-end object
  detection.
\newblock \emph{CoRR}, abs/2010.04159, 2020.

\bibitem[Wang et~al.(2021)Wang, Yu, Yu, Dai, Tsvetkov, and
  Cao]{wang_2021_simvlm}
Z.~Wang, J.~Yu, A.~W. Yu, Z.~Dai, Y.~Tsvetkov, and Y.~Cao.
\newblock Simvlm: Simple visual language model pretraining with weak
  supervision.
\newblock \emph{arXiv:2108.10904 [cs]}, 08 2021.

\bibitem[Yuhas et~al.(1989)Yuhas, Goldstein, and
  Sejnowski]{yuhas_1989_integration}
B.~Yuhas, M.~Goldstein, and T.~Sejnowski.
\newblock Integration of acoustic and visual speech signals using neural
  networks.
\newblock \emph{IEEE Communications Magazine}, 27:\penalty0 65--71, 11 1989.
\newblock \doi{10.1109/35.41402}.

\bibitem[Tiippana(2014)]{tiippana_2014_what}
K.~Tiippana.
\newblock What is the mcgurk effect?
\newblock \emph{Frontiers in Psychology}, 5, 07 2014.

\bibitem[McGurk and MacDonald(1976)]{McGurk1976HearingLA}
H.~McGurk and J.~MacDonald.
\newblock Hearing lips and seeing voices.
\newblock \emph{Nature}, 264:\penalty0 746--748, 1976.

\bibitem[Zhang et~al.(2021)Zhang, Li, Hu, Yang, Zhang, Wang, Choi, and
  Gao]{Zhang_2021_CVPR}
P.~Zhang, X.~Li, X.~Hu, J.~Yang, L.~Zhang, L.~Wang, Y.~Choi, and J.~Gao.
\newblock Vinvl: Revisiting visual representations in vision-language models.
\newblock In \emph{CVPR}, pages 5579--5588, 2021.

\bibitem[Wang et~al.(2020)Wang, Hu, Zhang, Li, Wang, Zhang, Gao, and
  Liu]{DBLP:journals/corr/abs-2012-06946}
J.~Wang, X.~Hu, P.~Zhang, X.~Li, L.~Wang, L.~Zhang, J.~Gao, and Z.~Liu.
\newblock Minivlm: {A} smaller and faster vision-language model.
\newblock \emph{CoRR}, abs/2012.06946, 2020.

\bibitem[Gothoskar et~al.(2020)Gothoskar, L{\'a}zaro-Gredilla, Agarwal,
  Bekiroglu, and George]{Gothoskar2020LearningAG}
N.~Gothoskar, M.~L{\'a}zaro-Gredilla, A.~Agarwal, Y.~Bekiroglu, and D.~George.
\newblock Learning a generative model for robot control using visual feedback.
\newblock \emph{ArXiv}, abs/2003.04474, 2020.

\bibitem[Ma et~al.(2022)Ma, Sodhani, Jayaraman, Bastani, Kumar, and
  Zhang]{Ma2022VIPTU}
Y.~J. Ma, S.~Sodhani, D.~Jayaraman, O.~Bastani, V.~Kumar, and A.~Zhang.
\newblock Vip: Towards universal visual reward and representation via
  value-implicit pre-training.
\newblock \emph{ArXiv}, abs/2210.00030, 2022.

\bibitem[Strokina et~al.(2022)Strokina, Yang, Pajarinen, Serbenyuk,
  K{\"a}m{\"a}r{\"a}inen, and Ghabcheloo]{Strokina2022VisualRF}
N.~Strokina, W.~Yang, J.~Pajarinen, N.~Serbenyuk, J.-K. K{\"a}m{\"a}r{\"a}inen,
  and R.~Ghabcheloo.
\newblock Visual rewards from observation for sequential tasks: Autonomous pile
  loading.
\newblock \emph{Frontiers in Robotics and AI}, 9, 2022.

\bibitem[Mohtasib et~al.(2021)Mohtasib, Neumann, and
  Cuay{\'a}huitl]{Mohtasib2021ASO}
A.~Mohtasib, G.~Neumann, and H.~Cuay{\'a}huitl.
\newblock A study on dense and sparse (visual) rewards in robot policy
  learning.
\newblock In \emph{TAROS}, 2021.

\end{thebibliography}
